\documentclass[journal,compsoc]{IEEEtran}
\ifCLASSOPTIONcompsoc
\usepackage[nocompress]{cite}
\else
\usepackage{cite}
\fi

\usepackage{multirow}
\usepackage{graphicx}
\usepackage{amsmath}
\usepackage{algorithm}
\usepackage{algorithmic}
\usepackage{amssymb}
\usepackage{hhline}
\usepackage{threeparttable}
\ifCLASSOPTIONcompsoc
\usepackage[caption=false,font=footnotesize,labelfont=sf,textfont=sf]{subfig}
\else
\usepackage[caption=false,font=footnotesize]{subfig}
\fi

\usepackage{url}
\usepackage{breakurl}
\usepackage[breaklinks]{hyperref}

\ifCLASSINFOpdf
\else
\fi
\begin{document}
\title{Fast-SNN: Fast Spiking Neural Network by Converting Quantized ANN}

\author{Yangfan~Hu,
	Qian~Zheng,~\IEEEmembership{Member,~IEEE},
	Xudong~Jiang,~\IEEEmembership{Fellow,~IEEE},
	and~Gang~Pan,~\IEEEmembership{Member,~IEEE}
	
	\IEEEcompsocitemizethanks{\IEEEcompsocthanksitem Y. Hu, Q. Zheng, G. Pan are with College of Computer Science and Technology, Zhejiang University, Hangzhou 310027, China. \protect\\
		E-mail:huyangfan@zju.edu.cn;~qianzheng@zju.edu.cn;~gpan@zju.edu.cn
		\IEEEcompsocthanksitem X. Jiang is with School of Electrical and Electronic Engineering, Nanyang Technological University, Singapore 639798, Singapore. \protect\\
		E-mail:exdjiang@ntu.edu.sg
		
	}
	\thanks{Manuscript received}}

\markboth{Journal of \LaTeX\ Class Files,~Vol.~14, No.~8, August~2015}%
{Shell \MakeLowercase{\textit{et al.}}: Bare Demo of IEEEtran.cls for Computer Society Journals}
%

\IEEEtitleabstractindextext{%
	\begin{abstract}
		Spiking neural networks (SNNs) have shown advantages in computation and energy efficiency over traditional artificial neural networks (ANNs) thanks to their event-driven representations. SNNs also replace weight multiplications in ANNs with additions, which are more energy-efficient and less computationally intensive. However, it remains a challenge to train deep SNNs due to the discrete spike function. A popular approach to circumvent this challenge is ANN-to-SNN conversion. However, due to the quantization error and accumulating error, it often requires lots of time steps (high inference latency) to achieve high performance, which negates SNN's advantages. To this end, this paper proposes Fast-SNN that achieves high performance with low latency. We demonstrate the equivalent mapping between temporal quantization in SNNs and spatial quantization in ANNs, based on which the minimization of the quantization error is transferred to quantized ANN training. With the minimization of the quantization error, we show that the sequential error is the primary cause of the accumulating error, which is addressed by introducing a signed IF neuron model and a layer-wise fine-tuning mechanism. Our method achieves state-of-the-art performance and low latency on various computer vision tasks, including image classification, object detection, and semantic segmentation. Codes are available at: \url{https://github.com/yangfan-hu/Fast-SNN}.
	\end{abstract}
	}

\maketitle

\IEEEdisplaynontitleabstractindextext

%
\IEEEpeerreviewmaketitle

\IEEEraisesectionheading{\section{Introduction}\label{sec:introduction}}
\IEEEPARstart{O}{ver} the last decade, deep artificial neural networks (ANNs) have made tremendous progress in various applications, including computer vision, natural language processing, speech recognition, etc. However, due to the increasing complexity in models and datasets, state-of-the-art ANNs require heavy memory and computational resources \cite{hubara2016binarized}. This situation prohibits the deployment of deep ANNs on resource-constrained environments (e.g., embedded systems or mobile devices). In contrast, the human brain can efficiently perform complex perceptual and cognitive tasks with a budget of approximately 20 watts \cite{roy2019towards}. Its remarkable capacities may be attributed to spike-based temporal processing that enables sparse and efficient information transfer in networks of biological neurons \cite{roy2019towards}. Inspired by biological neural networks, Maass \cite{maass1997networks} proposed a new class of neural networks, the spiking neural networks (SNNs). SNNs exchange information via spikes (binary events that are either 0 or 1) instead of continuous activation values in ANNs. An SNN unit (spiking neuron) only activates when it receives or emits a spike and remains dormant otherwise. Such event-driven, asynchronous characteristics of SNNs reduce energy consumption over time. In addition, SNNs use accumulate (AC) operations that are much less costly than the multiply-and-accumulate (MAC) operations in state-of-the-art deep ANNs. In the community of neuromorphic computing, researchers are developing neuromorphic computing platforms (e.g., TrueNorth \cite{merolla2014million}, Loihi \cite{davies2018loihi}) for SNN applications. These platforms, aiming at alleviating the von Neumann bottleneck with co-located memory and computation units, can perform SNN inference with low power consumption. Moreover, SNNs are inherently compatible with emerging event-based sensors (e.g., the dynamic vision sensor (DVS) \cite{lichtsteiner2008128}). 

However, the lack of efficient training algorithms obstructs deploying SNNs in real-time applications. Due to the discontinuous functionality of spiking neurons, gradient-descent backpropagation algorithms that have achieved great success in ANNs are not directly applicable to SNNs. Recently, researchers have made notable progress in training SNNs directly with backpropagation algorithms. They overcome the non-differentiability of the spike function by using surrogate gradients \cite{shrestha2018slayer,zenke2018superspike,wu2018spatio,gu2019stca,fang2021incorporating}. Then they apply backpropagation through time (BPTT) to optimize SNNs in a way similar to the backpropagation in ANNs. However, due to the sparsity of spike trains, directly training SNNs with BPTT is inefficient in both computation and memory with prevalent computing devices (e.g., GPUs) \cite{shrestha2018slayer,wu2021progressive}. Furthermore, the surrogate gradients would cause the vanishing or exploding gradient problem for deep networks, making direct training methods less effective for tasks of high complexity \cite{wu2021progressive}.

In contrast, rate-coded ANN-to-SNN conversion algorithms \cite{perez2013mapping,cao2015spiking,diehl2015fast,rueckauer2017conversion,hu2018spiking,Sengupta2019going,kim2020spiking,han2020rmp,deng2021optimal,yan2021near,li2021free,wu2021progressive} employ the same training procedures as ANNs, which benefit from the efficient computation of ANN training algorithms. Besides, by approximating the ANN activations with SNN firing rates, ANN-to-SNN conversion algorithms have achieved promising performance of SNNs on challenging tasks, including image classification \cite{rueckauer2017conversion} and object detection \cite{kim2020spiking}. Nevertheless, all existing methods suffer from quantization error and accumulating error \cite{rueckauer2017conversion} (see Section \ref{sec:conversion} for further discussion), resulting in performance degradation during conversion, especially when the latency is short. Although we can significantly mitigate the quantization error with an increasing number of discrete states (higher inference latency), it will unfavorably reduce the computation/energy efficiency of real-time applications. The growing inference latency will proportionally increase the number of operations and actual running time for a deployed SNN. 

To this end, this paper aims to build a Fast-SNN with competitive performance (i.e., comparable with ANNs) and low inference latency (i.e., 3, 7, 15) to conserve the SNN's advantages. Our basic idea is to reduce the quantization error and the accumulating error. The main contributions are:
\begin{itemize}
	\item \textbf{Quantization error minimization.} We show the equivalent mapping between temporal quantization in SNNs and spatial quantization in ANNs. Based on this mapping, we demonstrate that the quantization error could be minimized by the supervised training of quantized ANNs, which facilitates ANN-to-SNN conversion by finding the optimal clipping range and the novel distributions of weights and activations for each layer. Besides, we derivate the upper bound of inference latency that ensures the lossless conversion from quantized ANNs to SNNs.

	\item \textbf{Sequential error minimization.} To further boost the speed of SNNs and reduce the inference latency, we minimize the accumulating error. We show that the sequential error at each layer is the primary cause of the accumulating error when converting a quantized ANN to an SNN. Based on this observation, we propose a signed IF neuron to mitigate the impact from wrongly fired spikes to address the sequential error at each layer and propose a layer-wise fine-tuning mechanism to alleviate the accumulating sequential error across layers.
	
	\item \textbf{Deep models for various computer vision tasks.} Our method provides a promising solution to convert deep ANN models to SNN counterparts. It achieves state-of-the-art performance and low latency on various computer vision tasks, including image classification (accuracy: 71.31\%, time steps: 3, ImageNet), object detection (mAP: 73.43\%, time steps: 7, PASCAL VOC 2007), and semantic segmentation (mIoU: 69.7\%, time steps: 15, PASCAL VOC 2012).


\end{itemize}

\section{Related Work}\label{sec:related}
We present recent advances of two promising routes to build SNNs: direct training with surrogate gradients and ANN-to-SNN conversion. 

\subsection{Direct Training with Surrogate Gradients}\label{sec:direct}
Direct training methods view SNNs as special RNNs \cite{neftci2019surrogate} and employ BPTT to backward gradients through time. To address the discontinuous, non-differentiable nature of SNNs, researchers employ surrogate gradients \cite{lee2016training,shrestha2018slayer,zenke2018superspike,lee2020enabling} to approximate the derivative of the spiking function, which is a Dirac function.  In \cite{wu2018spatio}, Wu et al. first proposed a spatio-temporal backpropagation (STBP) framework to simultaneously consider the spatial and timing-dependent temporal domain during network training. Wu et al. further proposed to enhance STBP with a neuron normalization technique \cite{wu2019direct}. In \cite{gu2019stca}, Gu et al. proposed spatio-temporal credit assignment (STCA) for BPTT with a temporal based loss function. To develop a batch normalization method customized for BPTT, Kim et al. \cite{kim2021revisiting} proposed a batch normalize through time (BNTT) technique. Similarly, Zheng et al. \cite{zheng2021going} proposed a threshold-dependent batch normalization (tdBN) for STBP. In \cite{zhang2020temporal}, Zhang et al. proposed TSSL-BP to break down error backpropagation across two types of inter-neuron and intra-neuron dependencies, achieving low-latency SNNs. In \cite{rathi2021diet}, Rathi et al. proposed to optimize the leakage and threshold in the LIF neuron model. In \cite{kim2022neural}, Kim et al. proposed a Neural Architecture Search (NAS) approach to find better SNN architectures. For other perspectives on direct training methods, readers can refer to \cite{roy2019towards,eshraghian2021training,zenke2021brain,zenke2021visualizing}.

With these emerging techniques, direct training methods can build SNNs low latency and high accuracy \cite{zhang2020temporal,rathi2021diet,deng2022temporal}. Various SNN applications with direct training methods have also emerged \cite{venkatesha2021federated,kim2021optimizing,kim2022lottery,bhattacharjee2022mime,kim2021privatesnn,stewart2022meta,kim2022beyond}. However, due to the sparsity of spike trains, directly training SNNs with BPTT is inefficient in both computation and memory with prevalent computing devices (e.g., GPUs) \cite{shrestha2018slayer,wu2021progressive}. An SNN with $T$ time steps propagates $T$ times iteratively during forward and backward propagation. Compared with a full-precision ANN, it consumes more memory and requires about $T$ times more computation time. Furthermore, the surrogate gradients would cause the vanishing or exploding gradient problem for deep networks, making direct training methods less effective for tasks of high complexity \cite{wu2021progressive}. For these reasons, we focus on ANN-to-SNN conversion methods for building SNNs. It is worth noting that SNNs with $T=1$ can be viewed as a special case of binary neural networks (BNNs) \cite{hubara2016binarized}, making direct training and ANN-to-SNN conversion methods interchangeable. However, these SNNs suffer from a significant performance drop compared with their corresponding full-precision ANNs.   

\begin{table*}[t]
	\centering
	\caption{A summary of recent ANN-to-SNN conversion methods with respect to the minimization of the quantization error and accumulating error.}
	\label{tab:methods}
	\begin{tabular}{c|c|c|c|c|c}
		\hline\hline
		\multirow{4}{*}{Method} &\multicolumn{2}{|c|}{Quantization Error} & \multicolumn{2}{c|}{Accumulating Error}&\begin{tabular}{@{}c@{}}Time Step\end{tabular}\\\cline{2-5}
		&\begin{tabular}{@{}c@{}}Clipping \\ Range\\Optimization\end{tabular} &\begin{tabular}{@{}c@{}}Distribution of\\Weights/Activations\\Optimization \end{tabular}  &\begin{tabular}{@{}c@{}}Error Types \end{tabular}  &Minimization 
		&\begin{tabular}{@{}c@{}} Required for\\SOTA\\Performance\end{tabular}\\\cline{2-6}
		\hline	 
		Sengupta et al. 2019\cite{Sengupta2019going} &\multirow{3}{*}{Statistical Analysis} &\multirow{3}{*}{N/A} & \multirow{3}{*}{Quantization \& Sequential}  &\multirow{3}{*}{N/A} &\multirow{3}{*}{$\geq$2048}\\
		Kim et al. 2020 \cite{kim2020spiking} &&&&\\
		Han et al. 2020 \cite{han2020rmp}  &&&&\\\hline
		
		Yousefzadeh et al. 2019 \cite{yousefzadeh2019conversion} &N/A &N/A &Quantization \& Sequential & N/A &$\geq$800 \\\hline		
		
		Hu et al. 2018 \cite{hu2018spiking} &\multirow{2}{*}{Statistical Analysis} &\multirow{2}{*}{N/A} & \multirow{2}{*}{Quantization \& Sequential}  &\multirow{2}{*}{N/A} &\multirow{2}{*}{$\geq$16}\\
		Deng et al. 2021 \cite{deng2021optimal} &&&&\\\hline
		Yan et al. 2021 \cite{yan2021near} &\multirow{3}{*}{Statistical Analysis} &\multirow{3}{*}{Local Optimum} & \multirow{3}{*}{Quantization \& Sequential} &\multirow{3}{*}{Fine-tuning} &\multirow{3}{*}{$\geq$16}\\
		Li et al. 2021 \cite{li2021free} &&&&&\\
		Wu et al. 2021 \cite{wu2021progressive} &&&&&\\\hline

		Zou et al. 2020 \cite{zou2020novel} &Fixed Value &Local Optimum &Sequential &N/A &$\geq$4 \\\hline
		
		Ours &Learning Based &Global Optimum &Sequential &Fine-tuning&3\\\hline\hline
	\end{tabular}
\end{table*}

\subsection{ANN-to-SNN Conversion}\label{sec:conversion}
The first ANN-to-SNN conversion method by P\'{e}rez-Carrasco et al. \cite{perez2013mapping} maps ANNs of sigmoid neurons into SNNs of leaky integrate-and-fire (LIF) neurons by scaling weights of pre-trained ANNs. The scaling factor is determined by neuron parameters (e.g., threshold, leak rate, persistence). Although this method achieves promising results on two DVS tasks (human silhouette orientation and poker card symbol recognition), it does no apply to other tasks as its hyperparameters (neuron parameters) are determined manually. In \cite{cao2015spiking}, Cao et al. demonstrated that the rectified linear unit (ReLU) function is functionally equivalent to integrate-and-fire (IF) neuron, i.e., LIF neuron with no leaky factor nor refractory period. With only one hyperparameter: the firing threshold of spiking neurons, SNNs of IF neurons can approximate ANNs with ReLU activation, no bias term, and average pooling. With the great success of ReLU in ANNs, most existing ANN-to-SNN conversion methods follow the framework in \cite{cao2015spiking}. 

With the framework in \cite{cao2015spiking}, the quantization error \cite{rueckauer2017conversion}, usually observed as the over-/under-activation of spiking neurons compared with ANN activations \cite{diehl2015fast}, is recognized as the primary factor that obstructs lossless ANN-to-SNN conversion. In an SNN, the spiking functionality inherently quantizes the inputs (clipping the inputs to a range represented by temporally discrete states) and introduces the quantization error at each layer. To mitigate the quantization error, researchers proposed various normalization methods. Diehl et al. \cite{diehl2015fast} proposed to scale the weights by the maximum possible activation. In \cite{rueckauer2017conversion}, Rueckauer et al. improved the weight normalization \cite{diehl2015fast} by using the 99.9th percentile of activation instead of the maximum. In \cite{kim2020spiking}, Kim et al. proposed to apply channel-wise weight normalization to eliminate extremely small activations. In \cite{Sengupta2019going}, Sengupta et al. proposed threshold balancing, an equivalent alternative to weight normalization, to dynamically normalize SNNs at run time. Following \cite{Sengupta2019going}, Han et al. \cite{han2020rmp} proposed to scale the threshold by the fan-in and fan-out of the IF neuron. However, these methods only optimize the clipping range based on statistical analysis, leaving the distribution of weights/activations not optimized. Different from the above methods, Yousefzadeh et al. \cite{yousefzadeh2019conversion} proposed to address the over-activation problem with a signed neuron model that adapts the firing rate based on the total membrane charges, with a positive/negative firing threshold of +1/-1. However, it also leaves the distribution of weights/activations not optimized. In addition, as analyzed in \cite{rueckauer2017conversion}, the quantization error at each layer will contribute to the accumulating error that severely distorts the approximation between ANN activations and SNN firing rates at deep layers. Therefore, these methods typically employ shallow architectures and require a latency of hundreds or even thousands of time steps to achieve high accuracy.  

To address the accumulating error, researchers proposed to train a fitting ANN and apply statistical post-processing. In \cite{hu2018spiking}, Hu et al. employed ResNet, a robust deep architecture, for conversion. They devised a systematic approach to convert residual connections. They reported that the observed accumulating error in a residual architecture is lower than that in a plain architecture. Furthermore, they proposed to counter the accumulating error by increasing the firing rate of neurons at deeper layers based on the statistically estimated error. In \cite{deng2021optimal}, Deng et al. proposed to train ANNs using a capped ReLU, i.e., ReLU1 and ReLU2. Then they applied a scaling factor to normalize the firing thresholds by the maximum activation of capped ReLU function.

Different statistical post-processing, an approach that employs ANNs to adjust the distribution of weights/activations emerges. In \cite{zou2020novel}, Zou et al. proposed to employ a quantization function with fixed steps during ANN training to improve the mapping from ANNs to SNNs. In \cite{yan2021near}, Yan et al. proposed a framework to adjust the pre-trained ANNs with the knowledge of temporal quantization in SNNs. They introduced a residual term in ANNs to emulate the residual membrane potential in SNNs and reduce the quantization error. In \cite{li2021free}, Li et al. introduced layer-wise calibration to optimize the weights of SNNs. In \cite{wu2021progressive}, Wu et al. proposed a hybrid framework called progressive tandem learning to fine-tune the full-precision floating-point ANNs with the knowledge of temporal quantization. This framework achieves promising results with a latency of 16. However, it still significantly suffers from quantization error when the latency is low, e.g., 3 (see our results in Section). 

In Table \ref{tab:methods}, we summarize how each SNN method addresses the quantization error and accumulating error. By transferring the minimization of the quantization error to quantized ANN training, our method is the first to employ supervised training to optimize both the clipping range and distribution of weights/activations. It facilitates ANN-to-SNN conversion with a learnable clipping range and a global optimum\footnote{ANN quantization training naturally encourages or constraints weights/activations to gather around the target quantized values distributed in an optimized range \cite{li2020additive}, \cite{han2021improving}, \cite{lee2021cluster}. We consider their distributions globally/locally optimum if they are from an ANN trained with/without such a constraint.} for the distribution of weights/activations. In contrast, methods such as \cite{yan2021near,li2021free,wu2021progressive} can only achieve a local optimum as they first train a full-precision floating-point (FP32) model and then apply fine-tuning to the FP32 model. As for \cite{zou2020novel}, it can only achieve a local optimum as it does not optimize the clipping range during training. Thanks to transferring the quantization error to ANN training, our method simplifies the complexity of the accumulating error as it contains only the sequential error. With this promising solution, we further explore various computer vision tasks (including image classification, object detection, and semantic segmentation) in contrast to previous SNN methods that primarily focus on image classification.

\section{Fast-SNN}
\begin{figure}[t]
	\centering
	\includegraphics[width=\linewidth]{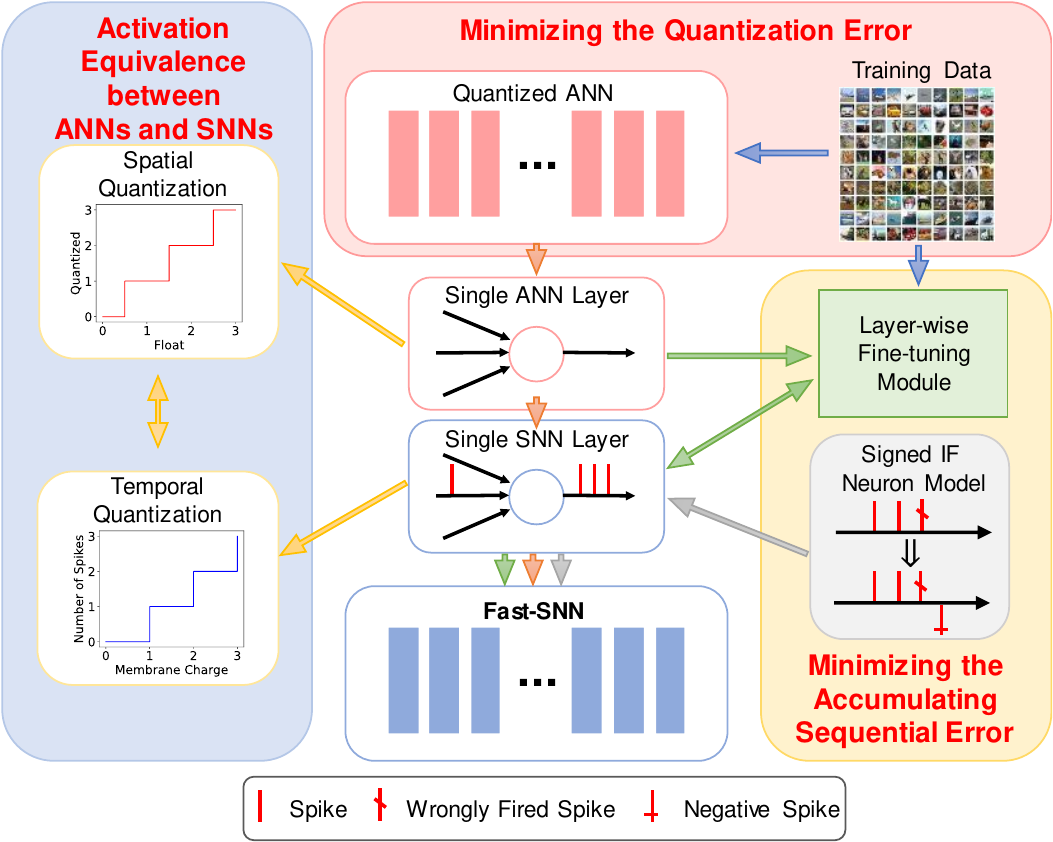}\\
	\caption{Overview of our conversion framework. We first minimize the quantization error during quantized ANN training. Then we provide an equivalence mapping between quantized ANN values and SNN firing rates. Finally, we address the accumulating error with a signed IF neuron model and a layer-wise fine-tuning module. Details of the layer-wise fine-tuning module can be found in Fig. \ref{fig:ft}.}\label{fig:overview}
\end{figure}

To build Fast-SNNs, we first analyze the equivalent mapping between \textit{spatial quantization} (mapping inputs to a set of discrete finite values) in ANNs and \textit{temporal quantization} (mapping inputs to a set of temporal events/spikes) in SNNs. This analysis reveals the activation equivalence between ANNs and SNNs, based on which we transfer the minimization of the quantization error to quantized ANN training. With the minimized quantization error, the accumulating sequential error, i.e., the combination of \textit{accumulating error} (an error caused by the mismatch in previous layers and grows with network propagation) and \textit{sequential error} (an error caused by the sequence and firing mechanism of spikes), becomes the primary factor that degrades SNN performance. To mitigate the sequential error at each layer, we introduce a signed IF neuron model. To alleviate the accumulating error across layers, we introduce a layer-wise fine-tuning mechanism to minimize the difference between SNN firing rates and ANN activations. Fig. \ref{fig:overview} illustrates the overview of our framework.

\subsection{Activation Equivalence between ANNs and SNNs}\label{sec:frame}
For a rate-coded spiking neuron $i$ at layer $l$, its spike count (number of output spikes) $N^l_i \in \left\{0, 1, \dots, T\right\}$, where $T$ is the length of spike trains. The spike count has $T+1$ discrete states (values). For a $b$-bit unsigned integer, it has $2^b$ discrete states (values): $\left\{0, 1, \dots, 2^b-1\right\}$. Our basic idea of the conversion from ANNs to SNNs is to map the integer activation of quantizaed ANNs $\left\{0, 1, \dots, 2^b-1\right\}$ to spike count  $\left\{0, 1, \dots, T\right\}$, i.e., set $T$ to $2^b-1$.

In the ANN domain, building ANNs with integer activations is naturally equivalent to compressing activations with the uniform quantization function that outputs uniformly distributed values. Such a function spatially discretizes a full-precision activation $x^l_i$ of neuron $i$ at layer $l$ in an ANN of ReLU activation into:     
\begin{equation}\label{eq:2}
	Q^l_i=\dfrac{s^l}{2^b-1}clip(round((2^b-1) \dfrac{x^l_i}{s^l}),0, 2^b-1),
\end{equation}
where $Q^l_i$ denotes the spatially quantized value, $b$ denotes the number of bits (precision), the number of states is $2^b-1$, $round(\cdot)$ denotes a rounding operator, $s^l$ denotes the clipping threshold that determines the clipping range of input $x^l_i$, $clip(x, min, max)$ is a clipping operator that saturates $x$ within range $[min, max]$. 

In the SNN domain, we consider the SNN model defined in \cite{rueckauer2017conversion} to facilitate conversion. This model employs direct coding (direct currents at the first layer) and IF (no leakage) neurons with reset-by-subtraction. At time step $t$, the total membrane charge $z^l_i(t)$ for neuron $i$ at layer $l$ is formulated as:
\begin{equation}\label{eq:3}
	z^l_i(t) = \sum\limits_{j=1}^{M^{l-1}}W^l_{ij}S^{l-1}_j(t) + b^l_i,
\end{equation}
where $M^{l-1}$ is the set of neurons at layer $l-1$, $W^l_{ij}$ is the weight of synaptic connection between neuron $i$ and $j$, $b^l_i$ is the bias term indicates a constant injecting current, $S^{l-1}_j(t)$ indicates an input spike from neuron $j$ at time $t$. The membrane equation of IF neuron is then defined as follows: 
\begin{equation}\label{eq:4}
	V^l_i(t) = V^l_i(t-1) + z^l_i(t) - \theta^l \Theta(V^l_i(t) - \theta^l),
\end{equation}
where $\theta^l$ denotes the firing threshold, $t$ denotes the $t$-th time step, $\Theta$ is a step function defined as:
\begin{equation}\label{eq:5}
	\Theta(x) =
	\begin{cases}
		1  & \text{if $x \geq 0$,}\\
		0  & \text{otherwise.}
	\end{cases}   
\end{equation}
Given a spike train of length $T$, the total input membrane charge over the whole time window $\widetilde{x}^l_i$ is defined as:
\begin{equation}\label{eq:6}
	\widetilde{x}^l_i = \mu^l + \sum_{t=1}^{T}z^l_i(t),
\end{equation}
where $\mu^l$ is the initial membrane charge. The spiking functionality of IF neurons inherently quantizes $\widetilde{x}^l_i$ into a quantized value represented by the firing rate $r^l_i(t)$: 
\begin{equation}\label{eq:7}
	\widetilde{Q}^l_i=r^l_i(t)=\dfrac{N^l_i}{T}=\dfrac{1}{T}clip(floor(\dfrac{\widetilde{x}^l_i}{\theta^l}),0,T),
\end{equation}
where $\widetilde{Q}^l_i$ denotes the temporally quantized value, $floor(\cdot)$ denotes a flooring operator. Since the 1st spiking layer (i.e., $l=1$) receives direct currents as inputs, we have 
\begin{equation}\label{eq:8}
	\widetilde{x}^1_i = T x^1_i.
\end{equation} 
Comparing Eq. \ref{eq:7} with Eq. \ref{eq:2}, we let $\mu^l = \theta^l/2$, $T = 2^b-1$, $\theta^l=s^l$. 
Since a flooring operator can be converted to a rounding operator 
\begin{equation}\label{eq:9}
	floor(x+0.5) = round(x),
\end{equation}
we rewrite Eq. \ref{eq:7} to 
\begin{equation}\label{eq:10}
	\widetilde{Q}^l_i=\dfrac{1}{T}clip(floor(\dfrac{Tx^l_i}{\theta^l}),0,T) =\dfrac{Q^l_i}{s^l}.
\end{equation}
Then we scale the weights in the following layer to $s^l W^{l+1}$, making an output spike equivalent to a continuous output of value $s^l$. We can rewrite the effectively quantized value from Eq. \ref{eq:10} to:  
\begin{equation}\label{eq:11}
	\widetilde{Q}^l_i=\dfrac{s^l}{T}clip(floor(\dfrac{Tx^l_i}{\theta^l}),0,T)= Q^l_i.
\end{equation}

\begin{figure}[t]
	\centering
	\includegraphics[width=.9\linewidth]{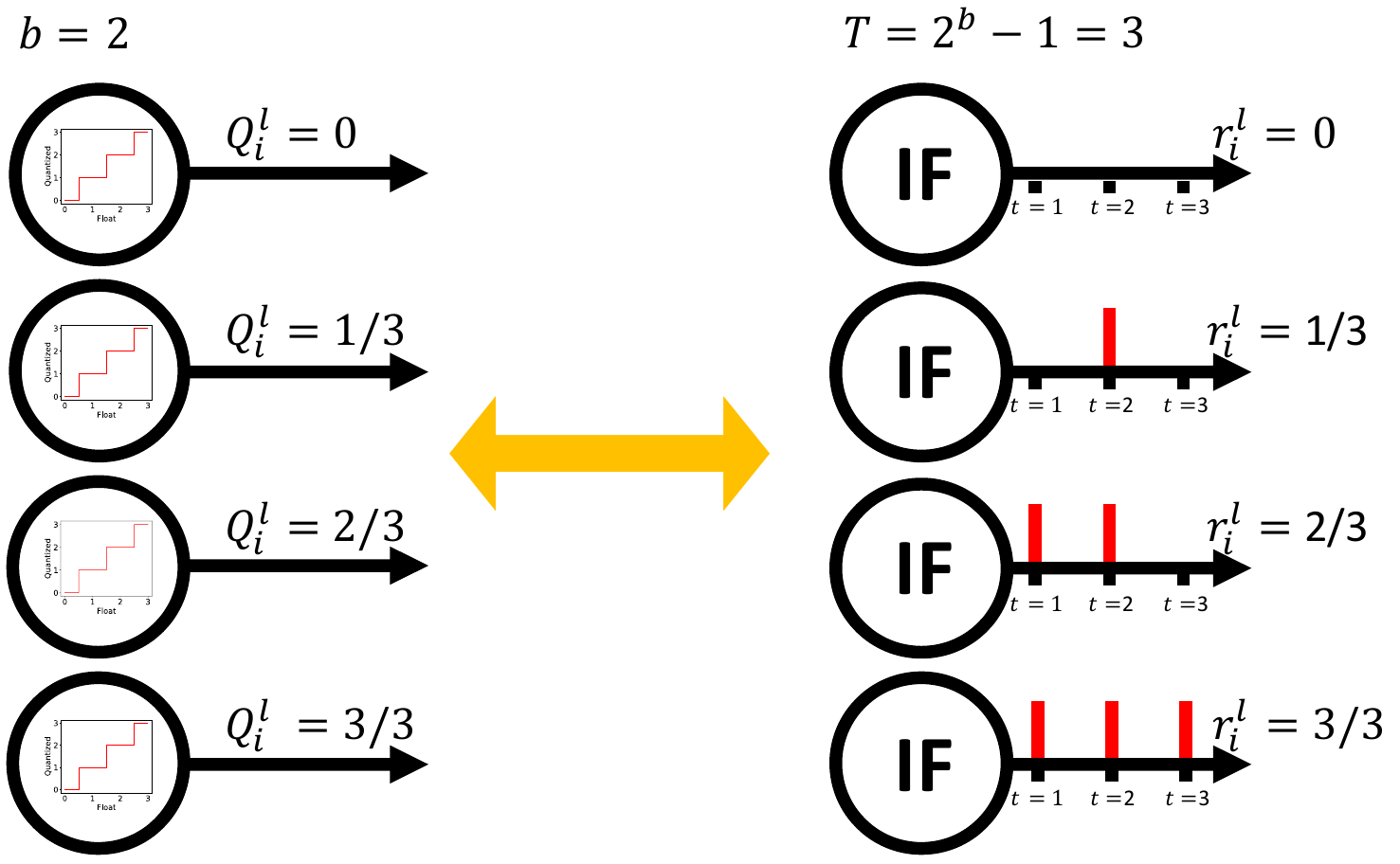}\\
	\caption{An example of how quantized values in ANNs are mapped to firing rates in SNNs. For a 2-bit ($b=2$) quantizer from Eq. \ref{eq:2} with $s^l=1$, it has four possible quantized values (e.g., 0, 1/3, 2/3, 3/3). Each quantized value is mapped to the spiking count (capped at $T=2^b-1$) in SNNs, resulting in four corresponding firing rates (e.g., 0, 1/3, 2/3, 3/3).}\label{fig:example}
\end{figure}

In Eq. \ref{eq:11}, our equivalent mapping between spatial quantization in ANNs and temporal quantization in SNNs derives the activation equivalence between ANNs and SNNs. In Fig. \ref{fig:example}, we present an example illustrating how quantized values in ANNs are mapped to firing rates in SNNs.

{\bf Latency upper bound derivation.} According to the equivalence between quantized ANNs and SNNs shown in Eq. \ref{eq:11}, the latency upper bound could be computed by forcing spiking neurons to start firing after receiving all possible spikes. In such a case, the sequential error of converting a quantized ANN to a rate-coded SNN is eliminated, and the firing rates in converted SNNs are identical to activations in quantized ANNs:
\begin{equation}\label{eq:x} 
	Latency = T \times L = (2^b-1) \times L.
\end{equation}
According to the latency bound in Eq. \ref{eq:x}, $b$ and $L$ are two factors that obstruct the fast inference of SNNs. To fully realize our Fast-SNN while maintaining network performance, we explore the minimization of $b$ in Section \ref{sec:quantization}, and reduce the impact from deep models (i.e., with a large $L$) in Section \ref{sec:sequence}.

\subsection{Quantization Error Minimization}\label{sec:quantization}
In Eq. \ref{eq:x}, the bit-precision of quantized ANNs determines $T$, the length of spike trains, and the latency. Since the number of operations in SNNs grows proportionally to $T$, a high bit-precision (e.g., $b=16$, $T=65535$) of quantized ANNs will negate SNN's advantages in computation/energy efficiency. The solution to reducing $T$ while retaining high accuracy is training low-precision networks (e.g., $b=2$, $T=3$) with minimized quantization error for conversion.            

Advances in ANN quantization \cite{hubara2016binarized} have shown that low-precision models can reach full-precision baseline accuracy with quantization-aware training \cite{li2020additive,han2021improving,lee2021cluster}. With Eq. \ref{eq:11}, we then equivalently convert the quantized ANNs to SNNs, eliminating the quantization error during conversion. Therefore, our framework fully inherits the advantages of quantization-aware training, minimizing the quantization error during training. On the one hand, we optimize network parameters with the knowledge of quantization, which minimizes the mismatch between the distribution of weights/activations and that of discrete quantization states \cite{li2020additive,han2021improving,lee2021cluster}. On the other hand, we optimize the clipping range for quantization by learning a clipping threshold during supervised training.

Our scheme of a learnable clipping range for quantization is quite different from existing SNN methods that clip the inputs to a range determined by a pre-defined scaling factor $\lambda$, e.g., finding the 99th or 99.9th percentile of all ReLU activation values \cite{rueckauer2017conversion,hu2018spiking,kim2020spiking,wu2021progressive}. Our method determines the clipping range using supervised training instead of statistical post-processing. Such difference brings three advantages: 
\begin{itemize}
	\item \textbf{Effectiveness.} Compared with a clipping threshold optimized during quantization-aware training in ANNs, $\lambda$ determined by the 99th or 99.9th percentile of all ReLU activation is not optimal. 	
	
	\item \textbf{Efficiency.} For each trained model, the normalization approach additionally calculates both the 99th and 99.9th percentile of ReLU activations for each ANN layer, respectively. Then it chooses the percentile (99th or 99.9th) that yields higher inference performance. The additional calculation would be notoriously costly if $\lambda$ is calculated from the whole training set. 
	
	\item \textbf{Stability.} For a challenging dataset such as ImageNet, it is impossible to calculate $\lambda$ from the whole training set. A solution is calculating $\lambda$ from a batch instead of the entire training set. However, the value of $\lambda$ will vary from batch to batch, and the performance of converted SNNs will fluctuate accordingly.  
\end{itemize}

\begin{figure}[t]
	\centering
	\includegraphics[width=1\linewidth]{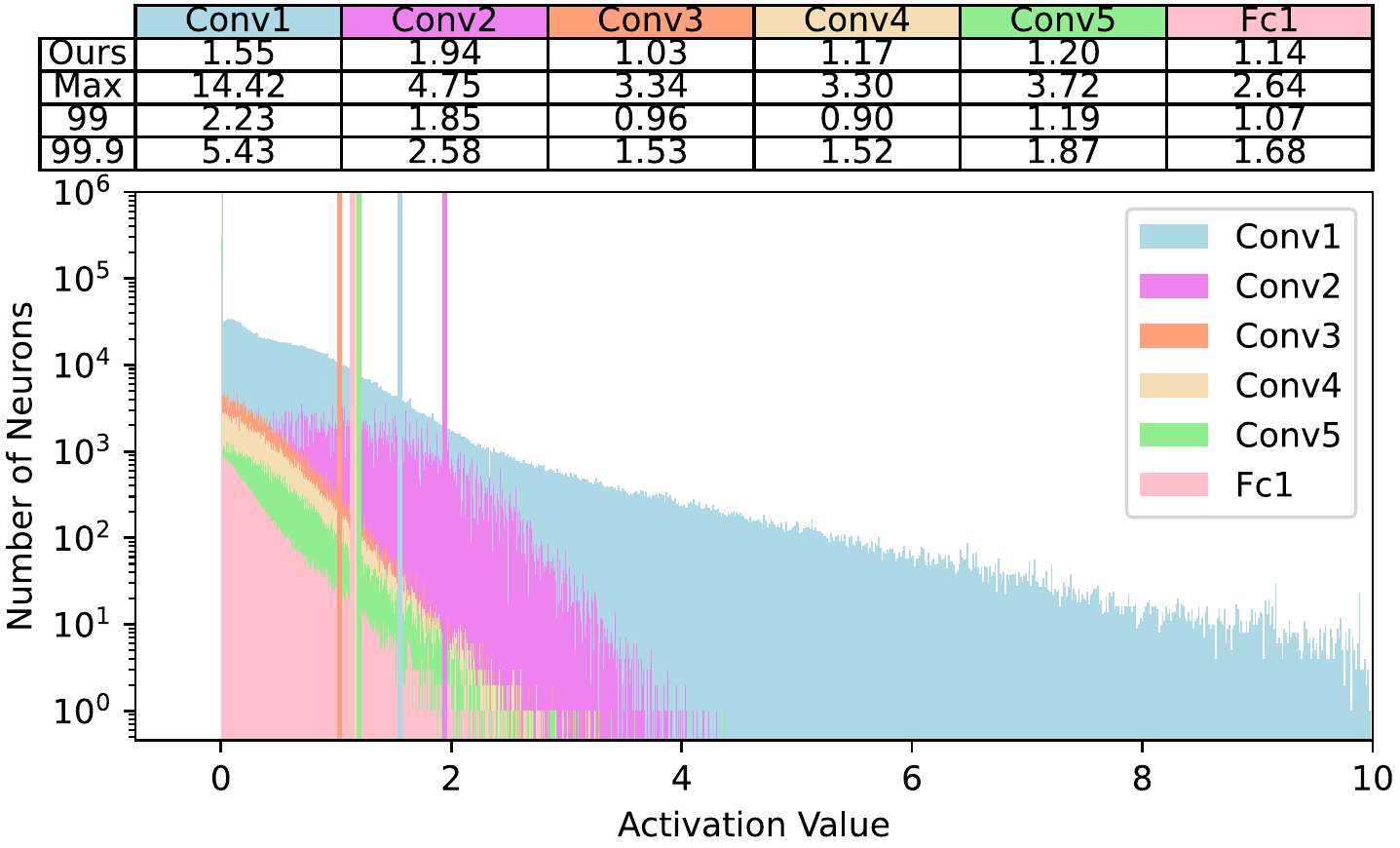}\\
	\caption{Distributions of the activations of a pre-trained 2-bit AlexNet using the first batch of the original training set. The batch size is set to 128. Solid lines denote our clipping thresholds. Values of the clipping threshold, max activation \cite{diehl2015fast}, 99th/99.9th percentile \cite{rueckauer2017conversion} of activations for each layer are listed within the table. Note that the max activation, 99th/99.9th percentile of activations are determined statistically from the same batch.}\label{fig:act}
\end{figure}

In Fig. \ref{fig:act}, we present the distributions of all ReLU activations for all layers of a pre-trained 2-bit AlexNet with the first batch of the original training set. As can be observed, the clipping threshold between our method and others is quite different. Our optimal clipping range brings a significant performance advantage, as shown in Section \ref{sec:threshold}.  

\subsection{Accumulating Error Minimization}\label{sec:sequence}
According to Eq. \ref{eq:x}, we can achieve lossless conversion from quantized ANNs to SNNs by enforcing a waiting period at each layer (spiking neurons can start firing after receiving all possible spikes for each layer). However, this scheme will prevent the practicability of our method on deep neural networks (i.e., when $L$ is large). A latency of $T\times L$ grows proportionally to the depth of the employed network, resulting in longer running time and potentially lower computation/energy efficiency in real-time applications (see our discussion in Section \ref{sec:discuss}). If we reduce the latency by removing the waiting period, it introduces the sequential error that degrades the performance of converted deep SNNs. 

\begin{figure}[t]
	\centering
	\includegraphics[width=.9\linewidth]{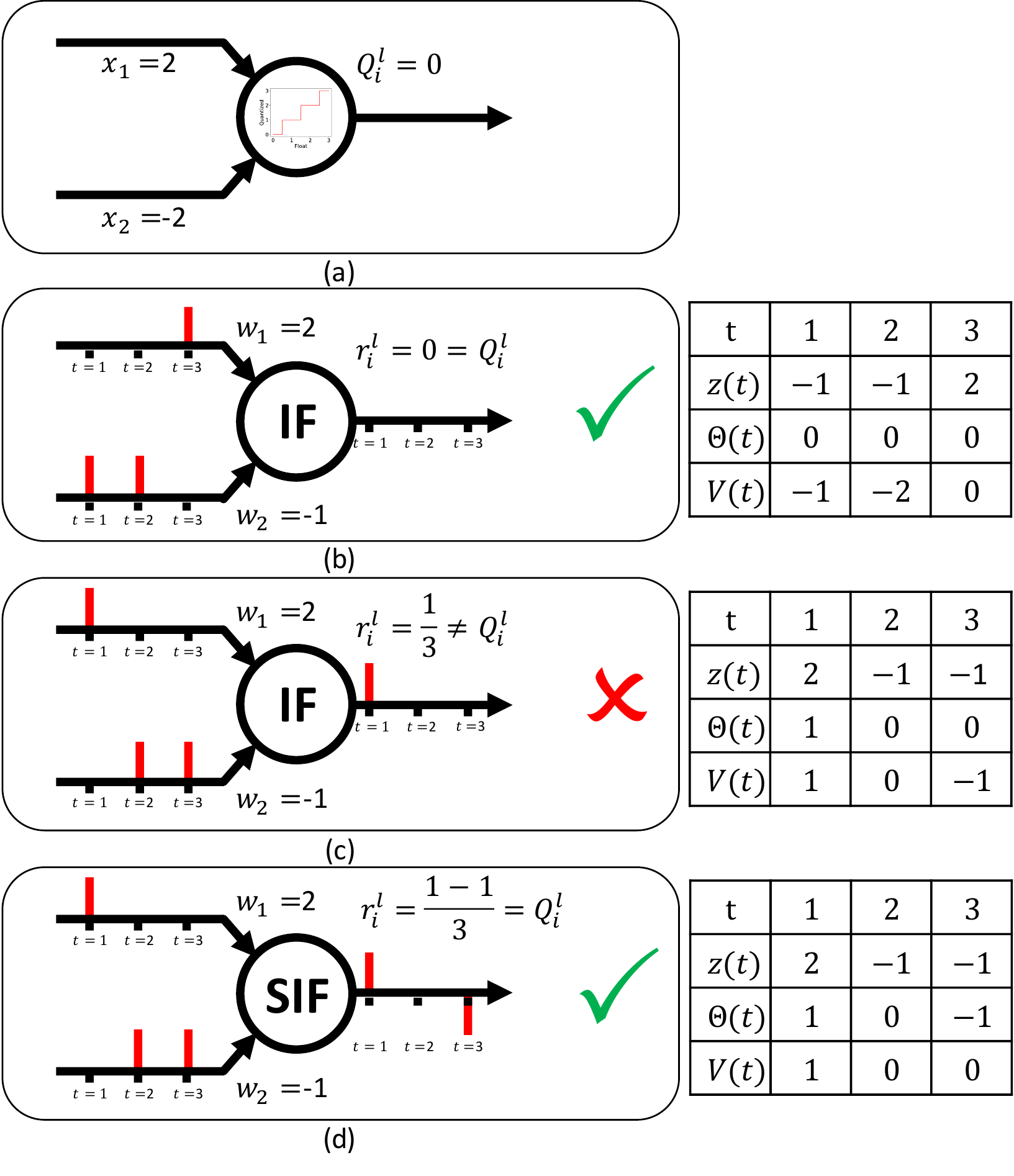}\\
	\caption{An example of how the sequence of spikes causes the sequential error. For simplicity, we set the firing threshold in SNNs to 1. SIF denotes our signed IF neuron. For each SNN neuron, we use a table to display the values of $z$, $\Theta$, $V$ for each time step t. (a) An ANN neuron receives two inputs: 2 and -2. Its output activation is 0. (b) An IF neuron receives three spike charges (-1, -1, 2) at $t=1, 2, 3$. Its output firing rate is equivalent to the ANN activation. (c) An IF neuron receives three spike charges (2, -1, -1) at $t=1, 2, 3$. However, it instantly fires a spike at $t=1$ since the membrane potential is greater than the firing threshold and outputs no events when $t=2, 3$, resulting in a firing rate that is not equivalent to the ANN activation. (d) An SIF neuron receives three spike charges (2, -1, -1) at $t=1, 2, 3$. Although it fires a spike at $t=1$, our SIF model outputs no events when $t=2$ as the incoming current cancels the residual membrane potential and fires a negative spike at $t=3$, resulting in a firing rate that is equivalent to the ANN activation.}\label{fig:sequential_error}
\end{figure}

In Fig. \ref{fig:sequential_error}, we demonstrate a case illustrating the cause and impact of the sequential error. Previous works of rate-coded SNNs (e.g., \cite{rueckauer2017conversion}, \cite{kim2020spiking}, \cite{han2020rmp}) seldom consider the sequential error because it has little impact on performance when $T$ is long enough (i.e., hundreds or thousands of time steps). That is, 
\begin{equation}
	\dfrac{N^l_i}{T} \approx \dfrac{N^l_i + 1}{T}, 
\end{equation}
where $N^l_i$ is the number of output spikes. However, when we reduce the latency $T$ to several time steps, the sequential error will significantly distort the approximation between ANN activations and SNN firing rates. Furthermore, the sequential error at each layer accumulates as the network propagates, causing significant deviation at deep layers.      

{\bf Signed IF neuron.}
To address the sequential error at each layer, we propose to cancel the wrongly fired spikes by introducing a signed IF neuron model. As for possible hardware implementation, neuromorphic hardware such as Loihi \cite{davies2018loihi} already supports signed spikes. In our signed IF neuron model, a neuron can only fire a negative spike if it reaches the negative firing threshold and has fired at least one positive spike. To restore the wrongly subtracted membrane potential, our model changes the reset mechanism for negative spikes to reset by adding the positive threshold (i.e., $\theta$). Then we rewrite the spiking function $\Theta(t)$ to
\begin{equation}\label{eq:12}
	\Theta(t) =
	\begin{cases}
		1  & \text{if $V^l_i(t) \geq \theta$,}\\
		-1 & \text{if $V^l_i(t) \leq \theta^{\prime}$ and $N^l_i(t)\geq 1$, }\\
		0  & \text{otherwise, no firing,}
	\end{cases}   
\end{equation}
where $\theta'$ is the negative threshold, $N^l_i(t)$ is the number of spikes neuron $i$ has fired at time step $t$. To boost the sensitivity of negative spikes, we set the negative threshold to a small negative value (empirically -1e-3). We then rewrite the membrane dynamic of IF neuron in Eq.\ref{eq:4} to:
\begin{equation}\label{eq:13}
	\begin{split}
		V^l_i(t) = &V^l_i(t-1) + z^l_i(t) - \theta^l \Theta(V^l_i(t) - \theta^l) \\
		&+ \theta^l \Theta(\theta^{\prime} - V^l_i(t))\Theta(N^l_i-1).\\
	\end{split}	
\end{equation}
With Eq.\ref{eq:12} and Eq.\ref{eq:13}, the IF neuron will fire a negative spike to cancel a wrongly fired spike and restore the membrane potential. Compared with our signed IF neuron model, the signed IF neuron model proposed by Kim et al. \cite{kim2020spiking} does not apply to our problem. It is designed to approximate negative outputs of the leaky ReLU function in ANNs and takes no consideration of the sequence of spikes in SNNs. In \cite{yousefzadeh2019conversion}, Yousefzadeh proposed a signed neuron model with a fixed positive/negative firing threshold of +1/-1, making it not sensitive to the sequential error.  

\begin{figure}[t]
	\centering
	\includegraphics[width=0.7\linewidth]{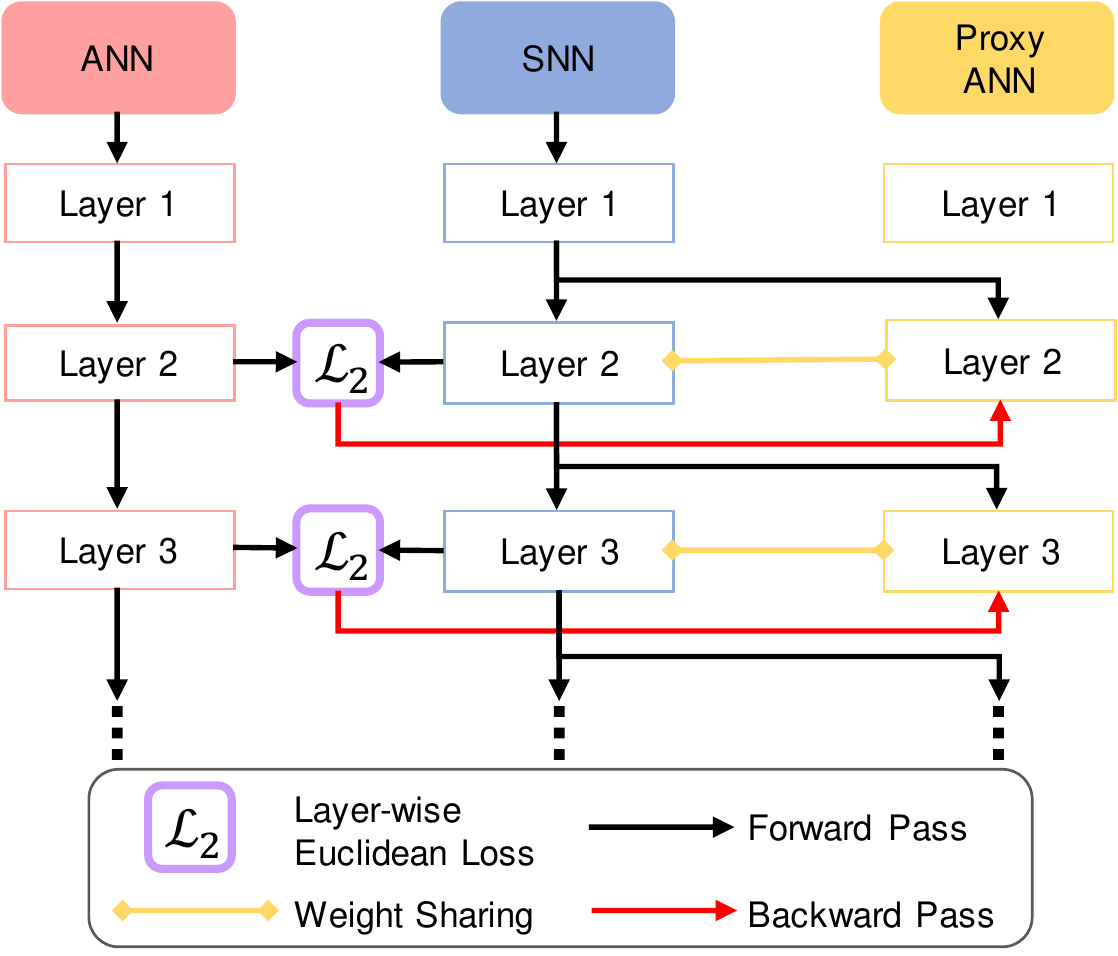}\\
	\caption{Layer-wise fine-tuning module. First, we obtain an SNN from a quantized ANN with $L$ layers. Then we build a proxy ANN that shares its parameters with the SNN. Starting from the 2nd layer (no sequential error in the 1st layer), layer $l$ of the proxy ANN receives the output map of firing rates from layer $l-1$ of the SNN as its input. Its output is set to the map of firing rates from layer $l$ of the SNN. We calculate the Euclidean loss between the output of layer $l$ in the proxy ANN and the reference (activations of layer $l$ in the quantized ANN). We then minimize the Euclidean loss by optimizing the parameters (weights and biases) of layer $l$ in the proxy ANN. The updated parameters in the proxy ANN are mapped back to the corresponding SNN layer. We repeat this process until reaching the final classification layer. We bypass the last layer as we directly use its membrane potential for classification. Please refer to Algorithm \ref{alg:algorithm} for more details.}\label{fig:ft}
\end{figure}

\begin{algorithm}[t]
	\caption{Minimizing the Accumulating Error}
	\label{alg:algorithm}
	\textbf{Input}: ANN $A$ , training set $\mathbb{X}$, latency $T$, network depth $L$\\
	\textbf{Output}: SNN model $S$
	\begin{algorithmic}[1] 
		\STATE //Network Initialization
		\STATE $S.load\_state(A.state\_dict())$
		\FOR{$layer \; l$ = $2$ to $L-1$}
		\STATE //Initialize a proxy ANN from $l$th layer of $A$
		\STATE $A_{p} = deepcopy(A.l)$
		\FOR{$x_i$ in $\mathbb{X}$}
		\STATE $A(x_i)$ //forward pre-trained ANN 
		\FOR{$t$ = $1$ to $T$}
		\STATE $S(x_i)$ //forward SNN
		\ENDFOR 
		\STATE $ref$ = $get\_activations(A,l)$ //reference
		\STATE $input$ = $get\_firing\_rates(S,l-1)$ //get inputs
		\STATE $output$ = $A_{p}(input)$ //forward proxy ANN 
		\STATE $output$ = $get\_firing\_rate(S,l)$ //update outputs
		\STATE $loss$ = $L2Loss(ref, output)$
		\STATE $optimizer$ step on $loss$
		\STATE $S.l.load\_state(A_{p}.state\_dict())$//update params
		\ENDFOR
		\ENDFOR
		\STATE \textbf{return} $S$
	\end{algorithmic}
\end{algorithm}

{\bf Layer-wise fine-tuning scheme.}
Although the modified neuron model narrows the sequential error at each layer, the accumulating error still distorts the SNN firing rates at deep layers and degrades network performance. According to our analysis, the firing rate maps in an SNN with a latency of $T \times L$ are identical to feature maps in quantized ANNs, indicating that the accumulating error in our framework contains only the accumulating sequential error. With this insight, we propose to minimize the accumulating error by minimizing the Euclidean distance between ANN activations (free from the sequential or accumulating error) and SNN firing rates at each layer with a framework illustrated in Fig. \ref{fig:ft}. To overcome the discontinuity in SNNs, we employ a proxy ANN that shares its parameters with the SNN to optimize network parameters (weights and biases). We present the pseudo-codes of our proposed layer-wise fine-tuning method in Algorithm \ref{alg:algorithm}. Compared with other fine-tuning methods \cite{yan2021near,li2021free,wu2021progressive}, our framework simplifies the optimization as it only needs to optimize the accumulating sequential error. Compared with layer-wise progressive tandem learning \cite{wu2021progressive} that fine-tunes all subsequent ANN layers together, our end-to-end fine-tuning mechanism yields less computation time and cost. Compared with layer-wise weight calibration \cite{li2021free}, our method incorporates the bias term (constant injecting current) during fine-tuning to learn a compensating membrane potential instead of calculating statistically.

\subsection{Implementation Details}
We perform all our experiments with PyTorch \cite{paszke2019pytorch}. To facilitate network training, we employ batch normalization layers \cite{ioffe2015batch} in our models to address the internal covariance shift problem with
\begin{equation}\label{eq:bn}
	\dfrac{x-\mu}{\sqrt{\sigma^2+\epsilon}}\gamma + \beta, 
\end{equation}
where $\mu$ is mini-batch mean, $\sigma^2$ is mini-batch variance, $\epsilon$ is a constant, $\gamma$ and $\beta$ are two learnable parameters. For hardware implementation, batch normalization can be incorporated into the firing threshold $\theta$ as
\begin{equation}\label{eq:bn2}
	\bar{\theta} = \dfrac{\theta-\beta}{\gamma}\sqrt{\sigma^2+\epsilon}+\mu. 
\end{equation}

To facilitate the training of deep networks, we employ shortcut connections from the residual learning framework \cite{he2016deep} to address the vanishing/exploding gradient problem. For hardware implementation, it only requires doubling pre-synaptic connections to receive inputs from the stacked layers and shortcuts, as the integration of spikes naturally performs addition operations.  

\begin{table}[t]
	\centering
	\caption{Accuracy (\%) of ANNs trained with different quantization precision on CIFAR-10. We denote the bit-precision of
		weights and activations by “W” and “A”, respectively. For the details of employed architectures, please refer to Section \ref{sec:setup1}.}
	\label{tab:anns_cifar10}
	\resizebox{\linewidth}{!}{
		\begin{tabular}{ccccccc}
			\hline 
			\quad 
			Network 
			&\begin{tabular}{@{}c@{}}Precision \\ W/A\end{tabular} 
			&\begin{tabular}{@{}c@{}}ANN \\ Acc.\end{tabular} 
			&\begin{tabular}{@{}c@{}}Precision \\ W/A\end{tabular} 
			&\begin{tabular}{@{}c@{}}ANN \\ Acc.\end{tabular} 			
			&\begin{tabular}{@{}c@{}}Precision \\ W/A\end{tabular} 
			&\begin{tabular}{@{}c@{}}ANN \\ Acc.\end{tabular} 	
			\\\hline
			AlexNet			&32/32 	&92.87 &3/3  &92.54 &2/2  &91.52\\
			VGG-11			&32/32	&93.60 &3/3  &93.71 &2/2  &93.06\\
			Resnet-20		&32/32 	&93.00 &3/3  &92.39 &2/2  &90.39\\
			Resnet-44		&32/32 	&94.17 &3/3  &93.41 &2/2  &91.57\\
			Resnet-56	 	&32/32	&94.10 &3/3  &93.66 &2/2  &91.68\\
			Resnet-18 		&32/32 	&95.85 &32/3 &95.62 &32/2 &95.51\\\hline
			
			\hline 
	\end{tabular}}
\end{table}

\begin{table}[t]
	\centering
	\caption{Accuracy (\%) of ANNs trained with different quantization precision on ImageNet. We denote the bit-precision of
		weights and activations by “W” and “A”, respectively.}
	\label{tab:anns_imagenet}
	\resizebox{\linewidth}{!}{
		\begin{tabular}{ccccccc}
			\hline 
			\quad 
			Network 
			&\begin{tabular}{@{}c@{}}Precision \\ W/A\end{tabular} 
			&\begin{tabular}{@{}c@{}}ANN \\ Acc.\end{tabular} 
			&\begin{tabular}{@{}c@{}}Precision \\ W/A\end{tabular} 
			&\begin{tabular}{@{}c@{}}ANN \\ Acc.\end{tabular} 			
			&\begin{tabular}{@{}c@{}}Precision \\ W/A\end{tabular} 
			&\begin{tabular}{@{}c@{}}ANN \\ Acc.\end{tabular} 	
			\\\hline
			AlexNet		&32/32 	&56.52 &32/3  &58.58 &32/2  &56.41\\
			VGG-16		&32/32	&73.36 &32/3  &73.02 &32/2  &71.91\\
			\hline 
	\end{tabular}}
\end{table}

To build quantized ANNs for conversion, we employ a state-of-the-art quantization method \cite{li2020additive} during training. To enable our ANN-to-SNN conversion method, we apply uniform quantization to activations. When exploring the building of low-precision SNNs, we quantize ANN weights with additive powers-of-two quantization instead of uniform quantization for better performance. In Table \ref{tab:anns_cifar10} and \ref{tab:anns_imagenet}, we demonstrate the classification accuracy of ANNs trained with different quantization precision on CIFAR-10 and ImageNet, respectively. On CIFAR-10, our ResNet-56 (ANN) with both weights and activations quantized to 3 bits achieves 93.66\% accuracy, which is close to the accuracy of full-precision ResNet-56 (ANN) implemented by us (94.10\%) or reported in \cite{he2016deep} (93.03\%). On ImageNet, our VGG-16 (ANN) with activations quantized to 3 bits achieves 73.02\% accuracy, only 0.34\% lower than the full-precision VGG-16 (ANN) from TorchVision \cite{torchvision2016}. Our VGG-16 (ANN) with activations quantized to 2 bits achieves 71.91\% accuracy on ImageNet, only 1.45\% lower than the full-precision VGG-16 (ANN) from TorchVision \cite{torchvision2016}. Based on the performance of these deep ANNs, we employ 3/2-bit ANNs for conversion in our experiments.  

Following previous works \cite{cao2015spiking,rueckauer2017conversion,deng2021optimal,li2021free,wu2019direct}, we employ the integrate-and-fire (IF) model as our spiking neuron model to better approximate ANN activations with SNN firing rates. For spike encoding, we employ the widely-used direct coding \cite{rueckauer2017conversion,deng2021optimal,li2021free,wu2021progressive,rathi2021diet}. The (positive) firing threshold $\theta$ is directly mapped from the scaling factor $s$ optimized in ANNs. As for the negative firing threshold $\theta^{\prime}$, we empirically set it to -1e-3 for all experiments.  

\section{Experiments on Image Classification}\label{sec:exp}

Image classification is a fundamental and heavily studied task in computer vision. It determines which objects are in an image or video. In the ANN domain, recent advances in image classification focus on training deep networks \cite{krizhevsky2012imagenet,simonyan2014very,he2016deep} and have achieved great success. In the SNN domain, image classification is also the most commonly used task for evaluation. To facilitate comparison with other SNN methods, we perform our primary experimental analysis in this section. 

\subsection{Experimental Setup}\label{sec:setup1}
\ \indent
\textbf{Datasets.} We perform image classification on two benchmark datasets: CIFAR-10 \cite{krizhevsky2009learning} and ImageNet \cite{ILSVRC15}. CIFAR-10 comprises 50,000 training images and 10,000 testing images in 10 classes. ImageNet comprises 1.2 million training images, 50,000 validation images, and 100,000 test images in 1,000 classes.

\textbf{Data preprocessing.} On CIFAR-10, we follow many works (e.g., \cite{he2016deep}) for data preprocessing, i.e., standardizing the data to have zero mean and unit variance, randomly taking a 32 $\times$ 32 crop from a padded 40 $\times$ 40 image (4 pixels padded on each side) or its horizontal flip. On ImageNet, we also follow existing works (e.g., \cite{desai2021virtex}) for data preprocessing, i.e., randomly cropping 10–100\% of the original image size with a random aspect ratio in (4/5, 5/4), resizing to 224 $\times$ 224, applying random flip and normalization by ImageNet color. For evaluation, we resize the input by its shorter edge to 256 pixels and take a 224 $\times$ 224 center crop.

\textbf{Network architecture.} 
On CIFAR-10, we train 32/3/2-bit AlexNet \cite{krizhevsky2012imagenet}, VGG-11 \cite{simonyan2014very}, and ResNet-18/20/44/56 \cite{he2016deep} for evaluation. For ResNet-20/44/56, we employ the original ResNet architectures defined in \cite{he2016deep}. To facilitate comparison with the ResNet-19 in \cite{deng2022temporal}, we employ a ResNet-18 similar to the ResNet-19 defined in \cite{deng2022temporal}. To explore the building of low-precision SNNs, we quantize both weights and activations for AlexNet, VGG-11, and ResNet-20/44/56 during ANN training. For ResNet-18, we quantize activations to enable conversion and keep full-precision weights for a fair comparison with other full-precision SNNs. On ImageNet, we train 3/2-bit AlexNet \cite{krizhevsky2012imagenet} and VGG-16 \cite{simonyan2014very} for evaluation. For AlexNet and VGG-16, we quantize activations to enable conversion and keep full-precision weights for a fair comparison with other full-precision SNNs. For full-precision models, we report the performance of pre-trained models from TorchVision \cite{torchvision2016}.    

\textbf{Training details.} We follow the training protocol defined in \cite{li2020additive}. We use stochastic gradient descent (SGD) with a momentum of 0.9. Table \ref{tab:lrwd} lists the weight decay and initial learning rate for different bit-precision on CIFAR-10 and ImageNet. On CIFAR-10, we divide the learning rate by 10 at the 150th, 225th epoch, and finish training at the 300th epoch. On ImageNet, we decrease the number of epochs in \cite{li2020additive} to 60, and divide the learning rate by 10 at the 20th, 40th, 50th epoch.  

\begin{table}[t]
	\centering
	\caption{Initial Learning Rate and Weight Decay. We denote the bit-precision of weights and activations by `W' and `A', respectively.}
	\label{tab:lrwd}
	\begin{tabular}{cccc}
		\hline 
		Dataset & Precision (W/A) & Learning Rate & Weight Decay\\\hline
		\multirow{4}{*}{CIFAR-10} & 32/32 & 0.1 & 5e-4\\
		& 4/4  & 4e-2& 1e-4\\
		& 3/3  & 4e-2& 1e-4\\
		& 2/2  & 4e-2& 3e-5\\\hline
		\multirow{2}{*}{ImageNet}& 32/3  & 1e-2& 1e-4\\
		& 32/2  & 1e-2& 1e-4\\
		\hline 
	\end{tabular}
\end{table}

\textbf{Evaluation metrics.}
In addition to the performance accuracy, we compare the computation/energy-efficiency of converted SNNs to their ANN counterparts by counting the number of operations during inference \cite{rueckauer2017conversion}. 

For an ANN, its number of operations is defined as:   
\begin{equation}
	Ops = \sum_{l=1}^{L} f^l_{in} M^l
\end{equation}
where $f_{in}$ denotes fan-in (number of incoming connections to a neuron), $L$ denotes number of layers, $M^l$ denotes the number of neurons at layer $l$. 

For an SNN, its number of operations is defined as the summation of all synaptic operations (number of membrane charges over time):
\begin{equation}
	Ops = \sum_{t=1}^{T}\sum_{l=1}^{L}\sum_{j=1}^{M^l}f^l_{out,j}s^l_j(t)
\end{equation}
where $f^l_{out,j}$ denotes fan-out of neuron $j$ at layer $l$ (number of output projections to neurons in the subsequent layer), $T$ denotes latency (number of time steps), $s^l_j(t)$ denotes number of spikes neuron $j$ at layer $l$ has fired at time step $t$.

\begin{table*}[t]
	\centering
	\caption{Performance comparison of different SNN methods on CIFAR-10 and ImageNet. Notions for different methods: Norm. is normalization \cite{Sengupta2019going,han2020rmp,deng2021optimal}, Cal. is calibration \cite{li2021free}, CQT is clamped and quantized training \cite{yan2021near}, PTL is progressive tandem learning \cite{wu2021progressive}, STDP-tdBN \cite{zheng2021going}, DIET-SNN \cite{rathi2021diet}, Grad. Re-wt. is gradient re-weighting \cite{deng2022temporal}. $\Delta$Acc. = SNN Acc. - ANN Acc. On ImageNet, bracketed numbers denote top-5 accuracy. We present the accuracy as a percent (\%). We denote the bit-precision of weights and activations by `W' and `A', respectively. Best and second best numbers are indicated by bold and underlined fonts, respectively.}
	\label{tab:compare}
	\resizebox{\linewidth}{!}{
		
		\begin{threeparttable}
			\begin{tabular}{cccccccccc}
				\hline \hline
				\quad &\textbf{Work} &\textbf{Architecture} &\textbf{Method} &\textbf{\begin{tabular}{@{}c@{}}Precision \\ (ANN W/A)\end{tabular}} &\textbf{\begin{tabular}{@{}c@{}}ANN \\ Acc.\end{tabular}} &\textbf{\begin{tabular}{@{}c@{}}Precision \\ (SNN W)\end{tabular}} &\textbf{\begin{tabular}{@{}c@{}}SNN \\ Acc.\end{tabular}} &\textbf{$\Delta$Acc.} &\textbf{\begin{tabular}{@{}c@{}}Time \\ Steps\end{tabular}}\\\hline\hline
				\multirow{20}{*}{\rotatebox{90}{\centering \textbf{CIFAR-10}}}
				&Sengupta et al. 2019 \cite{Sengupta2019going}	&VGG-16	&Norm.	&32/32	&91.70	&32	&91.55	&-0.15	&2500\\
				&Han et al. 2020 \cite{han2020rmp}	&VGG-16	&Norm.	&32/32	&93.63	&32	&93.63	&0	&2048\\
				&Deng et al. 2021 \cite{deng2021optimal}	&VGG-16	&Norm.	&32/32	&92.09	&32	&92.29	&+0.20 &16\\
				
				&Li et al. 2021 \cite{li2021free}	&VGG-16	&Cal.	&32/32	&95.72	&32	&93.71	&-2.01 &32\\
				
				&Yan et al. 2021 \cite{yan2021near}	&VGG-19	&CQT  &32/32 &93.60 &32	&93.44 &-0.06 &1000\\			
				&Yan et al. 2021 \cite{yan2021near}	&VGG-19	&CQT  &32/32 &93.60 &9	&93.43 &-0.07 &1000\\
				&Yan et al. 2021 \cite{yan2021near}	&VGG-19	&CQT  &32/32 &93.60 &8	&92.82 &-0.78 &1000\\
				&Wu et al. 2021 \cite{wu2021progressive}	&VGG-11		&PTL	&32/32	&90.59	&32	&91.24	&\underline{+0.65} &16\\
				&Wu et al. 2021 \cite{wu2021progressive}	&AlexNet	&PTL	&32/32	&89.59	&32	&90.86	&\textbf{+1.27} &16\\
				
				&Wu et al. 2021 \cite{wu2021progressive}	&AlexNet		&PTL	&-/-	&-	&8	&90.11	&- &16\\	
				&Wu et al. 2021 \cite{wu2021progressive}	&AlexNet		&PTL	&-/-	&-	&4	&89.48	&- &16\\
				
				&Zheng et al. 2021 \cite{zheng2021going}	&ResNet-19	&STBP-tdBN	&-/-	&-	&32	&93.16	&-	&6\\
				
				&Rathi et al. 2021 \cite{rathi2021diet}		&VGG-16		&DIET-SNN	&32/32	&93.72	&32	&92.70	&-1.02 &\underline{5}\\
				&Deng et al. 2022 \cite{deng2022temporal}	&VGG-16		&Grad. Re-wt.	&32/32	&94.97	&32	&94.50	&-0.47 &6\\	
	
				&Ours &AlexNet	&Fast-SNN	&3/3	&92.54	&3	&92.53	&-0.01	&7\\	
				&Ours &AlexNet	&Fast-SNN	&2/2	&91.52	&2	&91.63	&+0.11	&\textbf{3}\\
				&Ours &VGG-11 &Fast-SNN	&3/3	&93.71	&3	&93.72	&+0.01	&7\\	
				&Ours &VGG-11 &Fast-SNN	&2/2	&93.06	&2	&92.99	&-0.07	&\textbf{3}\\
				
				&Ours &ResNet-18 &Fast-SNN	&32/3	&95.62	&32	&\textbf{95.57}	&-0.05	&7\\	
				&Ours &ResNet-18 &Fast-SNN	&32/2	&95.51	&32	&\underline{95.42}	&-0.09	&\textbf{3}\\		
				
				\hline \hline
				\multirow{13}{*}{\rotatebox{90}{\centering \textbf{ImageNet}}}
				&Sengupta et al. 2019 \cite{Sengupta2019going}	&VGG-16	&Norm.	&32/32	&70.52 (89.39)	&32	&69.96 (89.01)	&-0.56 (-0.38)	&2500\\
				&Han et al. 2020 \cite{han2020rmp}	&VGG-16	&Norm.	&32/32	&73.49 (-)	&32	&\textbf{73.09} (-)	&-0.40 (-) &4096\\
				&Deng et al. 2021 \cite{deng2021optimal}	&VGG-16	&Norm.	&32/32	&72.40 (-)	&32	&55.80 (-)	&-16.60 (-) &16\\
				&Li et al. 2021 \cite{li2021free}	&VGG-16	&Cal.	&32/32	&75.36 (-)	&32	&63.64 (-)	&-11.72 (-) &32\\
				&Wu et al. 2021 \cite{wu2021progressive}	&AlexNet	&PTL	&32/32	&58.53 (81.07)	&32	&55.19 (78.41)	&-3.34 (-2.66) &16\\
				&Wu et al. 2021 \cite{wu2021progressive}	&VGG-16		&PTL	&32/32	&71.65 (90.37)	&32	&65.08 (85.25)	&-6.57 (-5.12) &16\\
				
				&Zheng et al. 2021 \cite{zheng2021going}	&ResNet-34	&STBP-tdBN	&-/-	&-	&32	&63.72 (-)	&-	&6\\
				&Rathi et al. 2021 \cite{rathi2021diet}		&VGG-16		&DIET-SNN	&32/32	&70.08 (-)	&32	&69.00 (-)	&-1.08 &\underline{5}\\	
				&Deng et al. 2022 \cite{deng2022temporal}	&ResNet-34\tnote{a}	&Grad. Re-wt.	&32/32	&-	&32	&64.79 (-)	&- &6\\	
	
				&Ours &AlexNet	&Fast-SNN	&32/3	&58.58 (80.57)	&32	&58.52 (80.59)	&\textbf{-0.06 (+0.02)}	&7\\
				&Ours &AlexNet	&Fast-SNN	&32/2	&56.41 (79.11)	&32	&56.34 (79.00)	&\underline{-0.07 (-0.11)}	&\textbf{3}\\
				&Ours &VGG-16 &Fast-SNN	&32/3	&73.02 (91.28)	&32	&\underline{72.95} (91.08)	&-0.07 (-0.20)	&7\\		
				&Ours &VGG-16 &Fast-SNN	&32/2	&71.91 (90.58)	&32	&71.31 (90.21)	&-0.60 (-0.37)	&\textbf{3}\\
				\hline \hline
			\end{tabular}
      		\begin{tablenotes}
				\footnotesize
				\item[a] For a fair comparison with other SNN methods, we report ResNet-34 in \cite{deng2022temporal} by the version that restricts spiking neurons to fire at most one spike at a single time step. 
			\end{tablenotes}
		\end{threeparttable}
	}
\end{table*}

\subsection{Overall Performance} \label{sec:cifar10}
In Table \ref{tab:compare}, we compare our method (Fast-SNN) with state-of-the-art SNN methods (including normalization \cite{Sengupta2019going,han2020rmp,deng2021optimal}, calibration \cite{li2021free}, clamped and quantized training \cite{yan2021near}, progressive tandem learning \cite{wu2021progressive}, STDP-tdBN \cite{zheng2021going}, DIET-SNN \cite{rathi2021diet}) using AlexNet \cite{krizhevsky2012imagenet}, VGG 11/16 \cite{simonyan2014very}, and ResNet-18 \cite{he2016deep}. All numbers for comparison are taken from corresponding papers, including the performance of pre-trained ANNs if provided. For a fair comparison with state-of-the-art conversion methods, we refer to calibration \cite{li2021free} and normalization \cite{han2020rmp} as our performance baseline on CIFAR-10 and ImageNet, respectively. We refer to progressive tandem learning \cite{wu2021progressive} as our latency baseline on both CIFAR-10 and ImageNet.

{\bf CIFAR-10.} Compared with the performance baseline \cite{li2021free}, our VGG-11 (latency is 7) achieves an accuracy slightly higher than their VGG-16 while using about 5$\times$ fewer time steps. Compared with the latency baseline \cite{wu2021progressive}, our AlexNet/VGG-11 (latency is 3) outperforms their AlexNet/VGG-11 by 0.77\%/1.75\% accuracy while using 5$\times$ fewer time steps. It is worth noting that the weight precision of our SNNs is 2-bit while their SNNs employ full-precision weights. Compared with their AlexNet with 4-bit weight precision, the accuracy of our AlexNet with 2-bit weight precision is 2.15\% higher than theirs. Compared with a state-of-the-art direct training method \cite{deng2022temporal}, our ResNet-18 (latency is 3) outperforms their ResNet-19 by 0.92\% accuracy while using half time steps. 

{\bf ImageNet.} Compared with the performance baseline \cite{han2020rmp}, our VGG-16 (latency is 7) achieves a top-1 accuracy only slightly lower than their VGG-16 while using about 600$\times$ fewer time steps. Compared with the latency baseline \cite{wu2021progressive}, our AlexNet/VGG-11 (latency is 3) outperforms their AlexNet/VGG-11 by 1.15\%(0.59\%)/6.23\%(4.96\%) on top-1 (top-5) accuracy while using 5$\times$ fewer time steps. Notably, our method achieves a small accuracy gap between the SNNs and their counterpart ANNs. With a latency of 7, the top-1 accuracy of our AlexNet/VGG-16 drops only 0.06\%/0.07\%. With a latency of 3, the top-1 accuracy of our AlexNet/VGG-16 drops 0.07\%/0.60\%. In contrast, the latency baseline \cite{wu2021progressive} reported a significant drop in accuracy (3.34\%/6.57\% for AlexNet/VGG-16 with a latency of 16) compared with pre-trained ANNs. Compared with a state-of-the-art direct training method \cite{rathi2021diet}, our VGG-16 (latency is 3) outperforms their VGG-16 by 2.31\% top-1 accuracy while using about half time steps.  

\subsection{Evaluation for Minimizing the Quantization Error}\label{sec:threshold}
To validate the efficacy of a learnable firing threshold, we train a 2-bit AlexNet/VGG-11 on CIFAR-10. Then we convert them to corresponding SNNs with different firing thresholds: the clipping threshold, the max activation, and the 99th/99.9th percentile of activations. We determine the max activation, 99th, and 99.9th percentile of activations by the first batch of the original training set with a batch size of 128. To compare the performance without sequential error, we set the latency to 3$\times$7/3$\times$11 for AlexNet/VGG-11 (spiking neurons start firing after receiving all possible spikes). To demonstrate the impact of the sequential error, we set the latency to 3. 

\begin{table}[t]
	\centering
	\caption{Accuracy (\%) of SNNs converted from 2-bit networks using different types of firing thresholds on CIFAR-10. We denote the maximum activation \cite{diehl2015fast} as Max, 99th/99.9th percentile \cite{rueckauer2017conversion} of activations as 99/99.9. We convert pre-trained ANNs directly to SNNs with no improvements applied. We denote the bit-precision of weights and activations by `W' and `A', respectively. Best numbers are indicated by the bold font.}
	\label{tab:vth_performance}
	\begin{tabular}{ccccc}
		\hline 
		Precision (W/A) &Network &Threshold &Acc. &Steps\\\hline
		2/2    &AlexNet(ANN) &-     &91.52	 & - \\	
		2/2	 &AlexNet(SNN) &Max	  &32.71	 & 3$\times$7\\
		2/2	 &AlexNet(SNN) &99.9  &72.29	 & 3$\times$7\\
		2/2	 &AlexNet(SNN) &99	  &89.16	 & 3$\times$7\\
		2/2	 &AlexNet(SNN) &Ours  &\textbf{91.52}	 & 3$\times$7\\\hline
		2/2	 &AlexNet(SNN) &Max	  &23.09	 & 3\\
		2/2	 &AlexNet(SNN) &99.9  &54.56	 & 3\\
		2/2	 &AlexNet(SNN) &99	  &81.66	 & 3\\
		2/2	 &AlexNet(SNN) &Ours  &\textbf{88.97}	 & 3\\\hline
		2/2    &VGG-11(ANN) &-     &93.06	 & - \\	
		2/2	 &VGG-11(SNN) &Max	 &21.25	 & 3$\times$11\\
		2/2	 &VGG-11(SNN) &99.9  &79.52	 & 3$\times$11\\
		2/2	 &VGG-11(SNN) &99	 &91.56  & 3$\times$11\\
		2/2	 &VGG-11(SNN) &Ours  &\textbf{93.06}	 & 3$\times$11\\\hline	
		2/2	 &VGG-11(SNN) &Max	 &24.04	 & 3\\
		2/2	 &VGG-11(SNN) &99.9  &45.56	 & 3\\
		2/2	 &VGG-11(SNN) &99	 &80.48	 & 3\\	
		2/2	 &VGG-11(SNN) &Ours  &\textbf{84.92}	 & 3\\					
		\hline 
	\end{tabular}
\end{table}

As shown in Table \ref{tab:vth_performance}, our learnable firing threshold achieves higher accuracy than all other types of firing thresholds under different latency configurations. Compared with normalization methods (statistical post-processing), our framework jointly optimizes the clipping threshold (firing threshold) at each layer, resulting in a clipping range that better fits the input data. It is worth noting that our SNNs using a latency of $(2^b-1)\times L$ (without the sequential error) achieve the same accuracy as their pre-trained ANNs. This result indicates no conversion error and validates the efficacy of our latency bound in Eq. \ref{eq:x}.  

\subsection{Evaluation for Minimizing the Accumulating Error}
To validate the efficacy of our signed IF neuron model and layer-wise fine-tuning mechanism, we stage by stage apply each piece of our method to SNNs with a latency of $2^b-1$. We train 3/2-bit AlexNet/VGG-11 on CIFAR-10 and AlexNet/VGG-16 on ImageNet with quantization-aware training. Then we convert the trained ANNs to SNNs. Here, we introduce a set of notions for SNNs. SNN$^\alpha$ stands for native SNNs directly converted from quantized ANNs. SNN$^\beta$ stands for SNN$^\alpha$ with our signed IF neuron model. SNN$^\gamma$ stands for SNN$^\beta$ with our layer-wise fine-tuning mechanism. 

\begin{table}[t]
	\centering
	\caption{Accuracy (\%) of SNN$^\alpha$ (native SNN), SNN$^\beta$ (SNN with our signed IF neuron model), SNN$^\gamma$ (SNN with both our signed IF neuron model and layer-wise fine-tuning) on CIFAR-10 and ImageNet. We denote the bit-precision of weights and activations by `W' and `A', respectively.}
	\label{tab:ablation}
	\resizebox{\linewidth}{!}{
	\begin{tabular}{cccccccc}
		\hline 
		\quad &\begin{tabular}{@{}c@{}}Precision \\ W/A\end{tabular} &Network 
		&\begin{tabular}{@{}c@{}}ANN \\ Acc.\end{tabular} 
		&\begin{tabular}{@{}c@{}}SNN$^\alpha$ \\ Acc.\end{tabular} 
		&\begin{tabular}{@{}c@{}}SNN$^\beta$ \\ Acc.\end{tabular} 
		&\begin{tabular}{@{}c@{}}SNN$^\gamma$ \\ Acc.\end{tabular}
		&\begin{tabular}{@{}c@{}}Time \\ Steps\end{tabular}\\\hline
		\multirow{4}{*}{\rotatebox{90}{\centering CIFAR-10}} 

		&3/3&AlexNet 	&92.54 &90.72 &92.46 &92.53 &7\\
		&2/2&AlexNet 	&91.52 &88.97 &91.37 &91.63 &3\\
		&3/3&VGG-11 	&93.71 &90.29 &93.46 &93.72 &7\\
		&2/2&VGG-11 	&93.06 &84.92 &92.80 &92.99 &3\\\hline

		\multirow{4}{*}{\rotatebox{90}{\centering ImageNet}}
		&32/3&AlexNet 	&58.58 &47.74 &58.35 &58.52 &7\\
		&32/2&AlexNet 	&56.41 &46.04 &55.93 &56.34 &3\\
		&32/3&VGG-16 	&73.02 &36.43 &72.89 &72.95 &7\\
		&32/2&VGG-16 	&71.91 &46.10 &71.10 &71.31 &3\\		
		\hline 
	\end{tabular}}
\end{table}

\begin{table}[t]
	\centering
	\caption{Accuracy (\%) of SNN$^\alpha$ (native SNN), SNN$^\beta$ (SNN with our signed IF neuron model), SNN$^\gamma$ (SNN with both our signed IF neuron model and layer-wise fine-tuning) on CIFAR-10. SNNs are converted from 3/2-bit ResNets with a latency of 7/3. $\Delta$Acc. = SNN$^\gamma$ Acc. - SNN$^\beta$ Acc.}
	\label{tab:deep}
	\begin{tabular}{ccccccc}
		\hline 
		\quad 
		&Network
		&\begin{tabular}{@{}c@{}}ANN \\ Acc.\end{tabular}
		&\begin{tabular}{@{}c@{}}SNN$^\alpha$ \\ Acc.\end{tabular}
		&\begin{tabular}{@{}c@{}}SNN$^\beta$ \\ Acc.\end{tabular}
		&\begin{tabular}{@{}c@{}}SNN$^\gamma$ \\ Acc.\end{tabular}
		&$\Delta$Acc.\\\hline
		
		\multirow{3}{*}{\rotatebox{90}{\centering 3-bit}}
		&ResNet-20 &92.39 &83.37 &91.33 &92.18 &+0.85\\
		&ResNet-44 &93.41 &61.06 &90.74 &92.62 &+1.88\\
		&ResNet-56 &93.66 &41.47 &88.55 &92.17 &+3.62\\
		
		\hline 
		\multirow{3}{*}{\rotatebox{90}{\centering 2-bit}}
		&ResNet-20 &90.39 &81.08 &88.81 &90.28 &+1.47\\
		&ResNet-44 &91.57 &59.84 &85.89 &89.59 &+3.70\\
		&ResNet-56 &91.68 &60.95 &81.25 &89.25 &+8.00\\
		
		\hline 
	\end{tabular}
\end{table}

As shown in Table \ref{tab:ablation}, our signed IF neuron model and layer-wise fine-tuning consistently improve the performance of converted SNNs. On CIFAR-10, SNN$^\beta$ with signed IF neuron model consistently improves the performance of SNN$^\alpha$ by at least 1.5\%. For the 2-bit VGG-11, the improvement achieves a wide margin of 7.88\%. With layer-wise fine-tuning applied, all our SNNs achieve almost lossless performance compared with the corresponding quantized ANNs. Recalling the full-precision ANNs results in Table 2 and Table 3, it is also notable our method achieves an accuracy comparable to full-precision ANN networks on CIFAR-10 and ImageNet with a latency of 7. Even with a latency of 3, the accuracy of SNN$^\gamma$ (AlexNet/VGG-11) on CIFAR-10 is only 1.24\%/0.61\% lower than full-precision ANNs. On ImageNet, the top-1 accuracy of SNN$^\gamma$ (AlexNet/VGG-16) with a latency of 3 is only 0.18\%/2.05\% lower than full-precision ANNs.    

In the above experiments, the improvement from our layer-wise fine-tuning mechanism is less significant, for the employed models (AlexNet, VGG-11, VGG-16) are relatively shallow (less than 20 layers). According to Eq. \ref{eq:x}, when $L$ is small, the impact of accumulating sequential error is limited, and SNN firing rates still approximate ANN activations. To further validate the efficacy of our layer-wise fine-tuning mechanism, we apply our layer-wise fine-tuning to deep residual networks (ResNets) \cite{he2016deep}. Kindly note that a deep architecture like ResNet could achieve high performance with fewer parameters and operations (e.g., on CIFAR-10, 32-bit AlexNet with an accuracy of 92.87\% uses 43$\times$ more parameters, 5$\times$ more operations than 32-bit ResNet-20 with an accuracy of 93.00\%). However, conventional SNNs may not benefit from deep architectures as the sequential error accumulates during network propagation. 

On CIFAR-10, we train 3/2-bit ResNet-20/44/56 and convert them to SNNs with a corresponding latency of 7/3. We measure the efficacy of our fine-tuning method by the difference between SNN$^\gamma$ accuracy and SNN$^\beta$ accuracy. In Table \ref{tab:deep}, our fine-tuning mechanism consistently improves the performance of all spiking ResNets. Notably, the accuracy gain from fine-tuning is bigger when the network is deeper or the precision is lower (lower latency). For the 2-bit ResNet-56, our fine-tuning method improves the accuracy of SNN$^\beta$ by a large margin of 8\%. This result coincides with our theoretical analysis that the accumulating sequential error grows proportional to network depth $L$. It also demonstrates the effectiveness of our method in re-approximating ANN activations with fine-tuned SNN firing rates.

\subsection{Computational Efficiency}\label{sec:efficiency}
In a neural network, the number of operations required for inference determines its computation/energy efficiency. In a rate-coded SNN, the number of operations grows proportionally to the latency. According to the analysis of the quantization error \cite{rueckauer2017conversion}, previous methods have to balance the trade-off between classification accuracy and inference latency. However, when the latency reaches a critical point, it invalidates SNN's advantages. For example, AlexNet (SNN), with a latency of 32, yields almost the same number of operations as AlexNet (ANN) on CIFAR-10. Therefore, we should also validate the computation/energy efficiency of SNNs concerning the latency required to achieve comparable accuracy to ANNs. Following \cite{rueckauer2017conversion}, we calculate the number of operations for our SNNs on CIFAR-10 with different inference latency. For comparison, we choose progressive tandem learning (PTL) \cite{wu2021progressive} that achieves the shortest latency among previous methods as our baseline. Then we reproduce AlexNet/VGG-11 from \cite{wu2021progressive} on CIFAR-10 using their public codes for both ANN training and SNN conversion with different latency configurations.    

\begin{figure}[t] 
	\centering
	\subfloat[\label{fig:a}]{%
		\includegraphics[width=.33\linewidth]{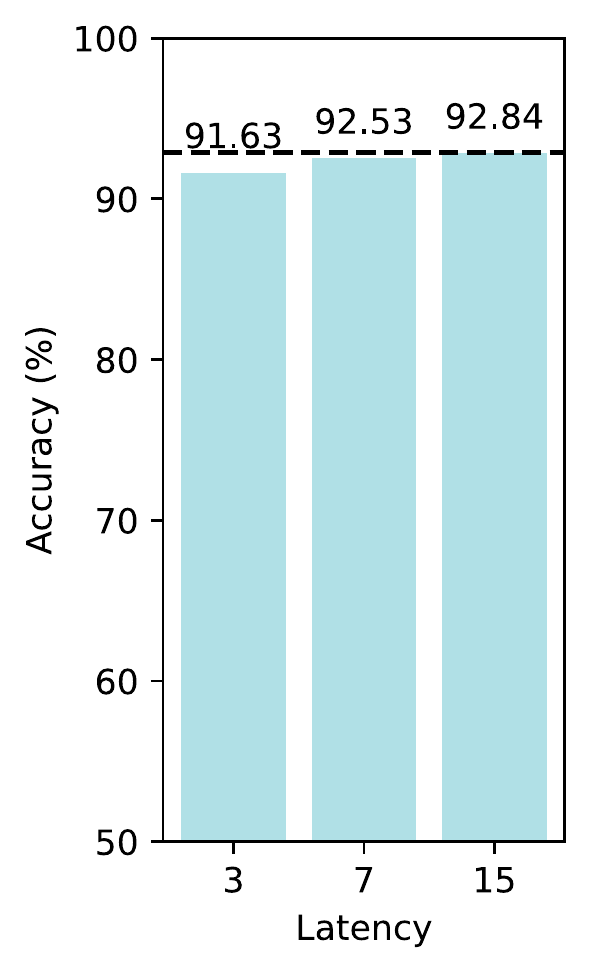}}
	\subfloat[\label{fig:b}]{%
		\includegraphics[width=.33\linewidth]{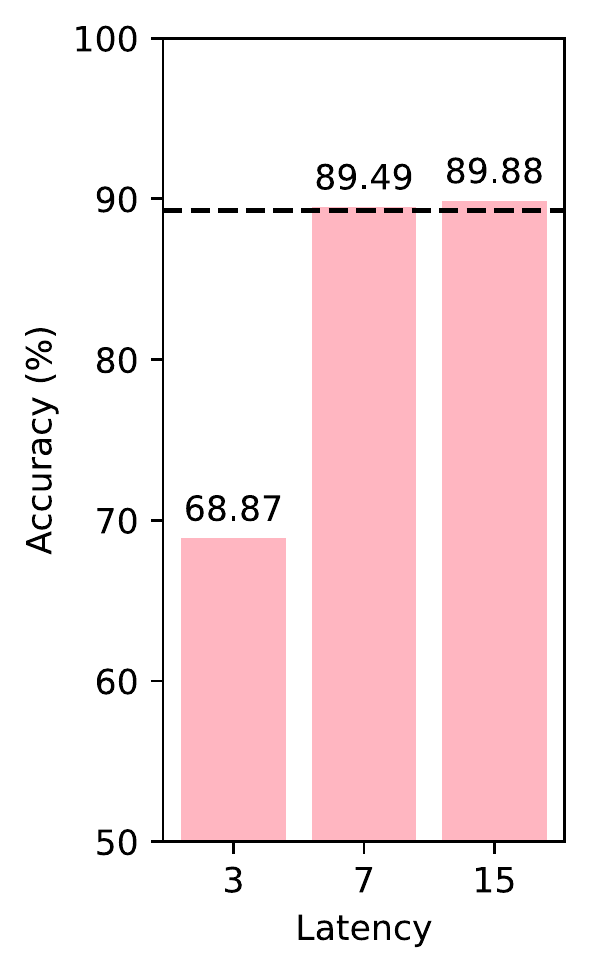}}
	\subfloat[\label{fig:c}]{%
		\includegraphics[width=.33\linewidth]{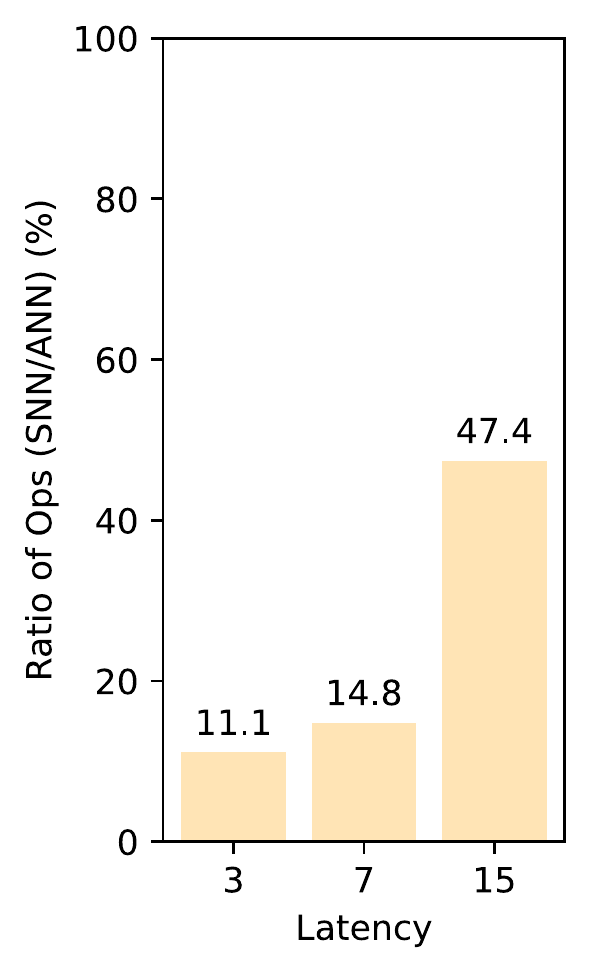}}	
	\\
	\subfloat[\label{fig:d}]{%
		\includegraphics[width=.33\linewidth]{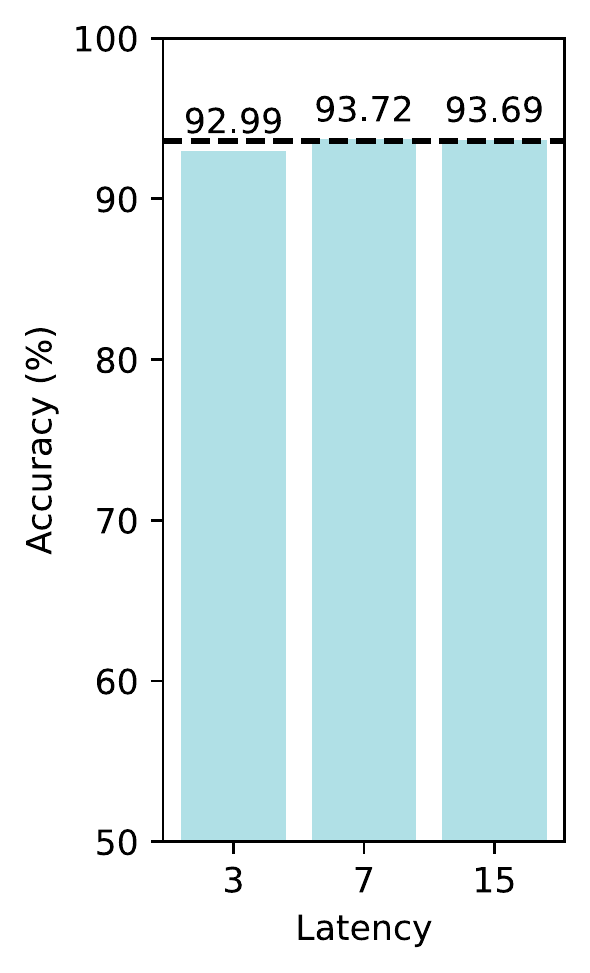}}
	\subfloat[\label{fig:e}]{%
		\includegraphics[width=.33\linewidth]{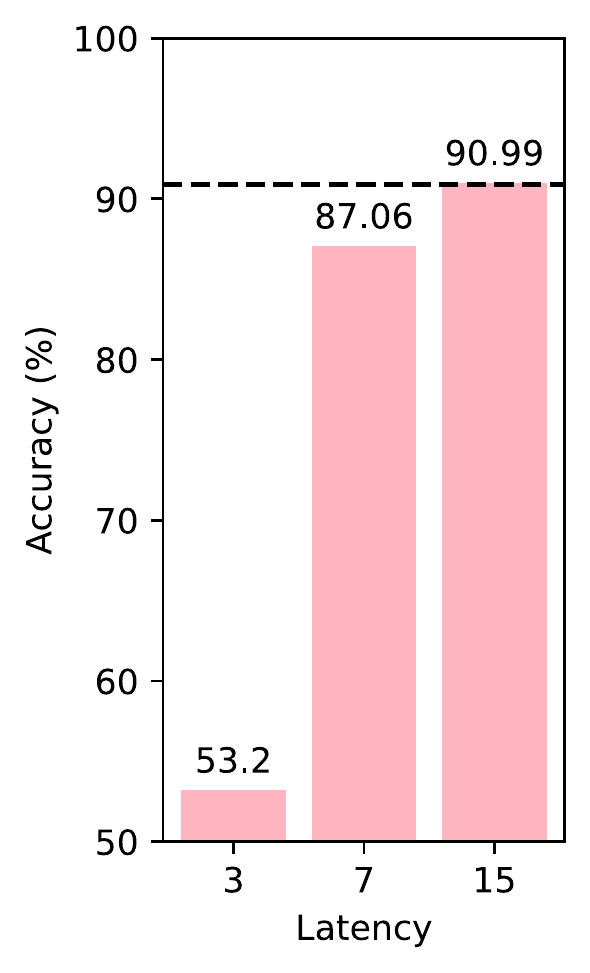}}
	\subfloat[\label{fig:f}]{%
		\includegraphics[width=.33\linewidth]{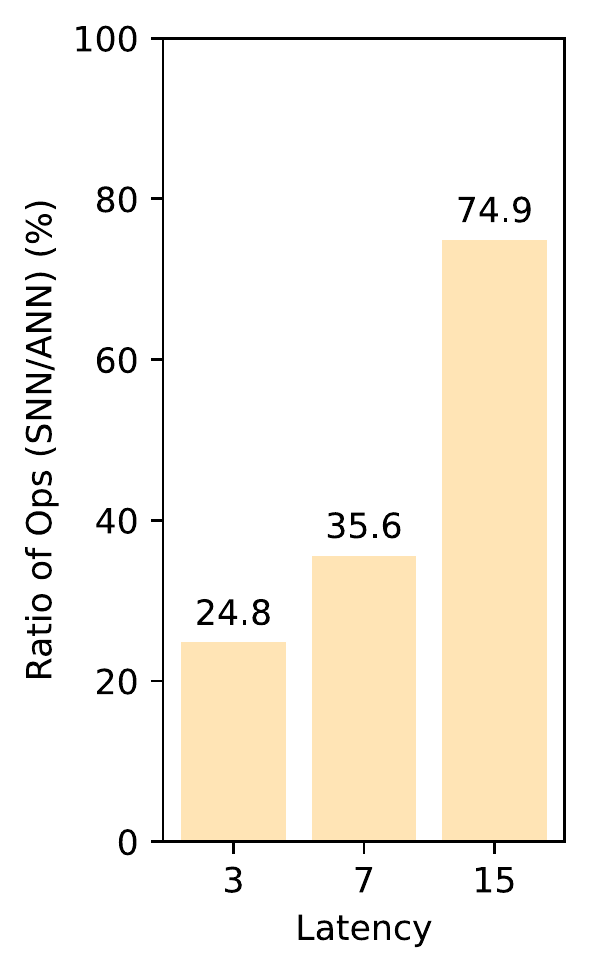}}		
	\caption{Performance comparison with PTL\cite{wu2021progressive} on AlexNet (top) and VGG-11 (bottom). Dashed lines indicate the performance of full-precision ANNs. (a) (d) Our performance. (b) (e) The performance of PTL [22]. (c) (f) Ratio of the number of operations (SNN to ANN), smaller is better.}
	\label{fig:ops}
\end{figure}

In Fig. \ref{fig:ops}, we show the performance comparison  ((a), (b), (d), (e)) and ratio of operations ((c), (f)) on CIFAR-10 regarding different latency. As can be observed, our method is much more computation/energy efficient compared with PTL \cite{wu2021progressive}. That is, the accuracy of our AlexNet/VGG-11 (latency is 3) is higher than that of PTL's \cite{wu2021progressive} AlexNet/VGG-11 (latency is 15) by 1.75\%/2.00\%. With a latency of 15, PTL [12] yields a ratio of operations of 47.4\%/74.9\%, while ours is 11.1\%/24.8\% with a latency of 3. This result indicates our AlexNet/VGG-11 is at least 4/3$\times$ more computation/energy efficient than PTL \cite{wu2021progressive}. 

\subsection{Discussion}\label{sec:discuss}
On CIFAR-10, we train the quantized ANNs with both weight and activation quantized. Therefore, the converted SNNs naturally inherit the weight precision of ANNs and are inherently compatible with neuromorphic hardware that supports low-bit integer weight precision. As shown in Table \ref{tab:compare}, \ref{tab:ablation}, and \ref{tab:deep}, our SNNs with quantized weights could achieve high performance with only a few time steps, making them friendly to real-time applications on low-precision neuromorphic hardware. 

In addition, a latency of 3 also improves the computation/energy efficiency of real-time SNN applications. Regarding the number of operations, our SNNs (latency is 3) are about 10$\times$ more efficient than their counterpart ANNs, not to mention operations in SNNs are more efficient than operations in ANNs. The deep ANNs are primarily composed of multiply-accumulate (MAC) operations that lead to high execution time, power consumption, and area overhead. In contrast, SNNs are composed of accumulate (AC) operations that are much less costly than MAC operations \cite{horowitz2014energy}. Recently, Arafa et al. \cite{arafa2020verified} reported that the energy consumption of an optimized 32-bit floating-point add instruction is about 10$\times$ lower than that of a multiply instruction under NVIDIA's Turing GPU architecture. 

Moreover, compared with the latency bound in Eq. \ref{eq:x}, our SNNs with a latency of $T$ improve the efficiency of real-time applications in terms of running time ($L \times$ faster). In practice, it also benefits energy consumption. Because even though the waiting period ideally does not introduce additional energy consumption as the number of operations remains unchanged, real-time applications still consume energy in a dormant state due to the restrictions in current neuromorphic devices. 


\begin{table*}[t]
	\centering
	\caption{Performance comparison for object detection task on PASCAL VOC 2007 and MS COCO 2017. `Norm.' denotes normalization. $\Delta$mAP = SNN mAP - ANN mAP. We present the mean AP as a percent (\%). We denote the bit-precision of weights and activations by `W' and `A', respectively. Best and second best numbers are indicated by bold and underlined fonts, respectively.}
	\label{tab:detect_results}
	\begin{tabular}{cccccccccc}
		\hline \hline
		\quad &\textbf{Work} &\textbf{Architecture} &\textbf{Method} &\textbf{\begin{tabular}{@{}c@{}}Precision \\ (ANN W/A)\end{tabular}} &\textbf{\begin{tabular}{@{}c@{}}ANN \\ mAP\end{tabular}} &\textbf{\begin{tabular}{@{}c@{}}Precision \\ (SNN W)\end{tabular}} &\textbf{\begin{tabular}{@{}c@{}}SNN \\ mAP\end{tabular}} &\textbf{$\Delta$mAP} &\textbf{\begin{tabular}{@{}c@{}}Time \\ Steps\end{tabular}}\\\hline\hline
		\multirow{8}{*}{\rotatebox{90}{\centering \textbf{PASCAL VOC}}}
		&Kim et al. 2020 \cite{kim2020spiking}	&Tiny YOLO	&Norm.	&32/32	&53.01	&32	&51.83	&-1.18	&8000\\
		&Kim et al. 2020 \cite{kim2020towards}	&Tiny YOLO	&Norm.	&32/32	&53.01	&32	&51.44	&-1.57	&5000\\
		
		&Ours &Tiny YOLO	&Fast-SNN	&32/4	&53.28	&32	&53.17 	&-0.11	&15\\
		&Ours &Tiny YOLO	&Fast-SNN	&32/3	&52.77	&32	&52.83	&\underline{+0.06}	&\underline{7}\\		
		&Ours &Tiny YOLO	&Fast-SNN	&32/2	&50.32	&32	&50.56	&\textbf{+0.24}	&\textbf{3}\\
		&Ours &YOLOv2(ResNet-34) &Fast-SNN	&32/4	&76.16	&32	&\textbf{76.05}	&-0.11	&15\\
		&Ours &YOLOv2(ResNet-34) &Fast-SNN	&32/3	&75.27  &32	&\underline{73.43}	&-1.84	&\underline{7}\\	
		&Ours &YOLOv2(ResNet-34) &Fast-SNN	&32/2	&73.57	&32	&68.57	&-5.00	&\textbf{3}\\
		
		\hline \hline
		\multirow{8}{*}{\rotatebox{90}{\centering \textbf{MS COCO}}}
		&Kim et al. 2020 \cite{kim2020spiking}	&Tiny YOLO	&Norm.	&32/32	&26.24	&32	&25.66	&-0.58	&8000\\
		&Kim et al. 2020 \cite{kim2020towards}	&Tiny YOLO	&Norm.	&32/32	&26.24	&32	&25.78	&-0.46	&5000\\
		&Ours &Tiny YOLO	&Fast-SNN	&32/4	&27.74	&32	&27.59	&\textbf{-0.15}	&15\\
		&Ours &Tiny YOLO	&Fast-SNN	&32/3	&26.84	&32	&26.49	&\underline{-0.35} 	&\underline{7}\\	
		&Ours &Tiny YOLO	&Fast-SNN	&32/2	&24.34	&32	&22.88	&-1.46	&\textbf{3}\\
		&Ours &YOLOv2(ResNet-34) &Fast-SNN	&32/4	&46.96	&32	&\textbf{46.40}	&-0.56	&15\\
		&Ours &YOLOv2(ResNet-34) &Fast-SNN	&32/3	&46.32	&32	&\underline{41.89}	&-4.43	&\underline{7}\\		
		&Ours &YOLOv2(ResNet-34) &Fast-SNN	&32/2	&43.33	&32	&33.84	&-9.49	&\textbf{3}\\
		\hline \hline
	\end{tabular}
\end{table*}

\begin{figure*}[t]
	\centering
	\includegraphics[width=\linewidth]{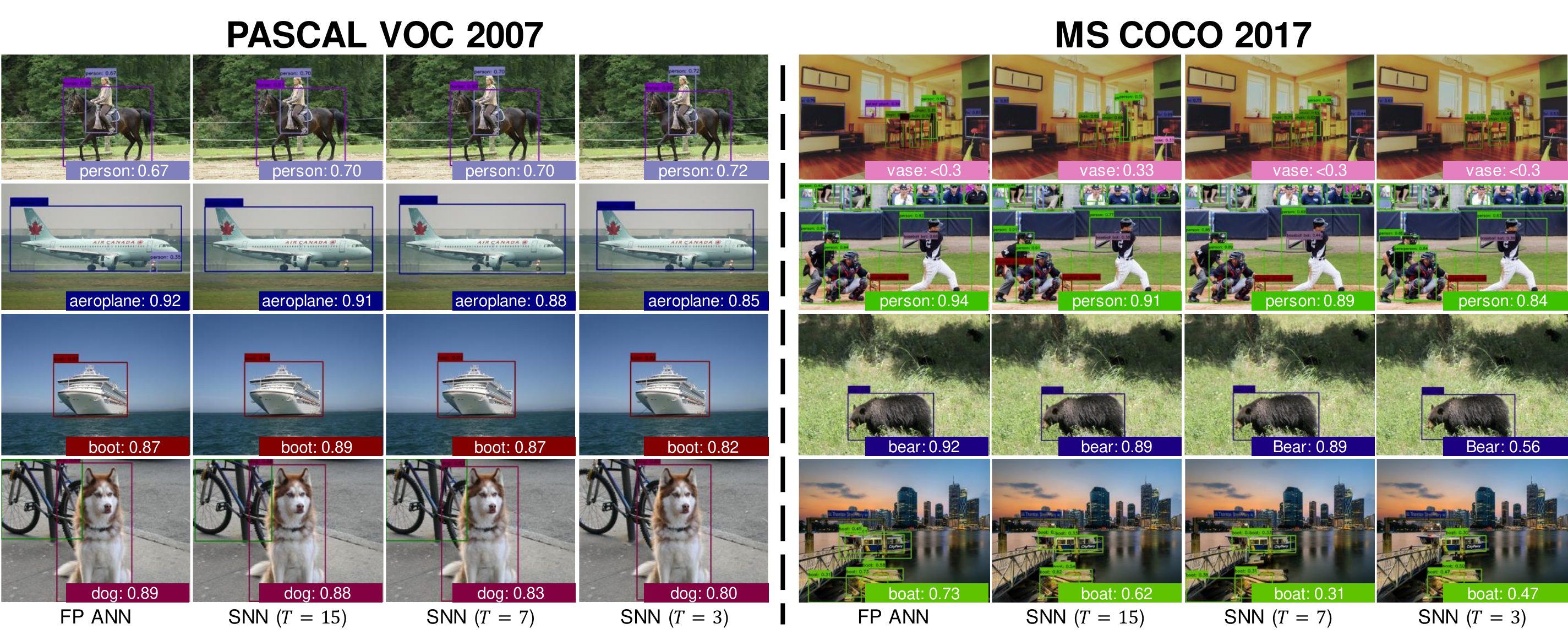}\\
	\caption{Visual quality comparison of object detection results on $test$ set of PASCAL VOC 2007 (left) and $val$ set of MS COCO 2017 (right) with the network architecture YOLOv2(ResNet-34).  From left to right: results from full-precision (FP) ANN models, our SNN with $T=15, 7, 3$, respectively.}\label{fig:detection}
\end{figure*}

\section{Experiments on Object Detection}

Object detection is a fundamental and heavily studied task in computer vision \cite{everingham2010pascal,lin2014microsoft}. It aims to recognize and locate every object in an image, typically with a bounding box. Thanks to the advances in deep learning, object detection has received significant improvements over recent years, especially with the application of Convolutional Neural Networks (CNNs). Currently, the main steam object detection algorithms fall into two lines of research. One line of research focuses on strong two-stage object detectors \cite{girshick2014rich,girshick2015fast,ren2015faster,he2017mask}. These Region-based CNN (R-CNN) \cite{girshick2014rich} algorithms first generate regions of interest (ROIs) with a region proposal network (RPN), then perform classification and bounding box regression. While being accurate, two-stage methods suffer from a slow inference speed. Another line of research focuses on one-stage object detectors \cite{redmon2016you,liu2016ssd}. These methods only use CNNs for feature extraction and directly predict the categories and positions of objects. With a balanced latency and accuracy, one-stage methods are widely used in real-time applications. Although heavily studied in the ANN domain, object detection requires further exploration in the SNN domain. An existing SNN object detector is the Spiking-YOLO \cite{kim2020spiking} and its improved version \cite{kim2020towards}. Spiking-YOLO employs an ANN-to-SNN conversion method with channel-wise normalization. However, Spiking-YOLO is unfriendly to real-time applications for requiring more than 5,000 time steps during inference. Here, we explore our Fast-SNN for object detection using the YOLO framework \cite{redmon2017yolo9000} for a fair comparison. 

\subsection{Experimental Setup}
\ \indent
\textbf{Datasets.} We perform object detection task on two datasets: PASCAL VOC \cite{everingham2010pascal} and MS COCO 2017 \cite{lin2014microsoft}. The PASCAL VOC dataset contains 20 object categories. Specifically, the train/val/test data in VOC 2007 contains 24,640 annotated objects in 9,963 images. For VOC 2012, the train/val data contains 27,450 annotated objects in 11,530 images. For MS COCO 2017 dataset, it contains 80 object categories. It has 886,284 annotated objects spread in 118,287 training images and 5,000 validation images. Following a common protocol \cite{ren2015faster,liu2016ssd,redmon2017yolo9000} on PASCAL VOC, the training data is the union of VOC 2007 trainval and VOC 2012 trainval datasets. As for testing, we use the VOC 2007 test dataset. 

\textbf{Data preprocessing.} We use a similar data augmentation to YOLO \cite{redmon2017yolo9000} and SSD \cite{liu2016ssd} with random crops, color shifting, etc.

\textbf{Network architecture.} Following \cite{kim2020spiking}, we employ a simple but efficient version of YOLO, the Tiny YOLO \cite{redmon2017yolo9000} for evaluation. We modify Tiny YOLO for ANN-to-SNN conversion by replacing all leakyReLU with ReLU. We also remove max pooling layers by incorporating the downsampling operations into convolution layers. To further explore object detection with deep SNNs, we also evaluate a YOLOv2 \cite{redmon2017yolo9000} with a backbone of ResNet-34. In the remainder of this paper, we refer this architecture as YOLOv2(ResNet-34). For PASCAL VOC 2007, we predict 5 boxes with 5 coordinates each and 20 classes per box, resulting in 125 filters. For MS COCO 2017, we predict 5 boxes with 5 coordinates each and 80 classes per box, resulting in 425 filters.

\textbf{Training details.} On PASCAL VOC and MS COCO, we fine-tune the models initialized from pre-trained ImageNet models to straightly adapt them to the object detection task. For Tiny YOLO, we initialize the backbone network from a model trained with the protocol in Section \ref{sec:setup1} on ImageNet. For YOLOv2(ResNet-34), we directly initialize the backbone from the pre-trained ResNet-34 in TorchVision \cite{torchvision2016}. Then we fine-tune the initialized models for 250 epochs using the standard SGD optimizer on PASCAL VOC 2007 and MS COCO 2017. For the first two epochs we slowly raise the learning rate from 0 to 1e-3. Then we continue training with a learning rate of 1e-3 and divide the learning rate by 10 at the the 150th, 200th epoch.  

\textbf{Evaluation metrics.} 
The performance is measured in terms of mean average precision (mAP). On PASCAL VOC, we use the definitions from VOC 2007 to calculate average precision (AP) and report the mAP over 20 object categories. On MS COCO, we follow MS COCO protocol to calculate AP and report the mAP over 80 object categories. For a fair comparison with Spiking-YOLO \cite{kim2020spiking}, we report the mAPs at IoU = 0.5 on both PASCAL VOC and MS COCO.  

\subsection{Overall Performance}
We summarize and compare the performance in Table \ref{tab:detect_results}. We include the Spiking-YOLO by Kim et al. \cite{kim2020spiking} and its improved version \cite{kim2020towards} for comparison. All numbers are taken from corresponding papers. On PASCAL VOC 2007, our Tiny YOLO (latency is 7) outperforms \cite{kim2020towards} (Tiny YOLO) by 1.39\% mean AP performance while using about $714 \times$ fewer time steps. Furthermore, our Tiny-YOLO achieves almost lossless conversion for all latency configurations (3, 7, 15). Thanks to the shallow architecture of Tiny YOLO, our converted Tiny YOLO even outperforms its counterpart ANN when the latency is 3/7. To further explore the capacity of deep SNNs in object detection, we apply our method to the more challenging YOLOv2(ResNet-34) architecture. Compared with \cite{kim2020towards}, our YOLOv2(ResNet-34) (latency is 15/7/3) achieves 24.61\%/21.99\%/17.13\% higher mean AP performance while using $333/714/1,667 \times$ fewer time steps. On MS COCO 2017, our Tiny YOLO (latency is 7) outperforms \cite{kim2020towards} (Tiny YOLO) by 0.71\% mean AP performance while using about $714 \times$ fewer time steps. Our YOLOv2(ResNet-34) (latency is 15/7/3) outperforms \cite{kim2020towards} by 20.62\%/16.11\%/7.67\% mean AP performance while using $333/714/1,667 \times$ fewer time steps. We further provide visual results of our YOLOv2(ResNet-34) in Fig. \ref{fig:detection}. As shown in the figure, our SNNs are able to detect objects at a degree close to the full-precision ANN. Our SNN with a latency of 15 can detect objects (e.g., vase) that the full-precision ANN fails to detect.   

\section{Experiments on Semantic Segmentation}
Semantic segmentation is a fundamental and heavily studied task in computer vision \cite{everingham2010pascal,lin2014microsoft}. It aims to predict the object class of every pixel in an image or give it a `background' status if not in listed classes. In recent years, semantic segmentation with deep learning has achieved great success. The fully convolutional network (FCN) \cite{long2015fully} that regards semantic segmentation as a dense per-pixel classification problem has been the basis of semantic segmentation with CNNs. To preserve image details, SegNet \cite{badrinarayanan2017segnet} employs an encoder-decoder structure, U-Net \cite{ronneberger2015u} introduces a skip connection between the downsampling and up-sampling paths, and RefineNet \cite{lin2017refinenet} presents multi-path refinement to exploit fine-grained low-level features. To capture the contextual information at multiple scales, Deeplab \cite{chen2017deeplab} introduces Atrous Spatial Pyramid Pooling (ASPP), and PSPNet \cite{zhao2017pyramid} performs spatial pyramid pooling at different scales. Although heavily studied in the ANN domain, semantic segmentation is scarcely explored in the SNN domain. An existing work by Kim et al. \cite{kim2022beyond} employs SNNs directly trained with surrogate gradients for semantic segmentation. However, this method suffers from a significant performance drop compared with ANNs. Here, we explore our Fast-SNN for semantic segmentation using the Deeplab framework \cite{chen2017deeplab} for a fair comparison.

\begin{table*}[t]
	\centering
	\caption{Performance comparison for semantic segmentation task on PASCAL VOC 2012 and MS COCO 2017. `DT' denotes direct training. $\Delta$mIoU = SNN mIoU - ANN mIoU. We present the mean IoU as a percent (\%). We denote the bit-precision of weights and activations by `W' and `A', respectively. Best and second best numbers are indicated by bold and underlined fonts, respectively.}
	\label{tab:seg_results}
	\begin{tabular}{cccccccccc}
		\hline \hline
		\quad &\textbf{Work} &\textbf{Architecture} &\textbf{Method} &\textbf{\begin{tabular}{@{}c@{}}Precision \\ (ANN W/A)\end{tabular}} &\textbf{\begin{tabular}{@{}c@{}}ANN \\ mIoU\end{tabular}} &\textbf{\begin{tabular}{@{}c@{}}Precision \\ (SNN W)\end{tabular}} &\textbf{\begin{tabular}{@{}c@{}}SNN \\ mIoU\end{tabular}} &\textbf{$\Delta$mIoU} &\textbf{\begin{tabular}{@{}c@{}}Time \\ Steps\end{tabular}}\\\hline\hline
		\multirow{7}{*}{\rotatebox{90}{\centering \textbf{PASCAL VOC}}}
		&Kim et al. 2021 \cite{kim2022beyond}	&VGG-9	&DT	&32/32	&34.7	&32	&22.3	&-12.4	&20\\
		&Ours &VGG-9	&Fast-SNN	&32/4	&44.38	&32	&44.55 &\textbf{+0.17} &15\\
		&Ours &VGG-9	&Fast-SNN	&32/3	&45.94	&32	&44.98	&\underline{-0.96} &\underline{7}\\		
		&Ours &VGG-9	&Fast-SNN	&32/2	&45.41	&32	&43.86	&-1.55	&\textbf{3}\\
		
		&Ours &ResNet-34 + ASPP &Fast-SNN	&32/4	&71.97	&32	&\textbf{69.7}	&-2.27	&15\\
		&Ours &ResNet-34 + ASPP &Fast-SNN	&32/3	&70.31  &32	&\underline{59.81}	&-10.50	&\underline{7}\\	
		&Ours &ResNet-34 + ASPP &Fast-SNN	&32/2	&41.45	&32	&17.28	&-24.17	&\textbf{3}\\
		
		\hline \hline
		\multirow{6}{*}{\rotatebox{90}{\centering \textbf{MS COCO}}}
		&Ours &VGG-9	&Fast-SNN	&32/4	&31.63	&32	&31.14	&\textbf{-0.49} &15\\
		&Ours &VGG-9	&Fast-SNN	&32/3	&31.50	&32	&30.46	&\underline{-1.04} &\underline{7}\\	
		&Ours &VGG-9	&Fast-SNN	&32/2	&30.20	&32	&28.35	&-1.85	&\textbf{3}\\
		&Ours &ResNet-34 + ASPP	&Fast-SNN	&32/4	&52.22	&32	&\textbf{50.24}	&-1.98	&15\\
		&Ours &ResNet-34 + ASPP &Fast-SNN	&32/3	&52.15	&32	&\underline{47.03}	&-5.12	&\underline{7}\\		
		&Ours &ResNet-34 + ASPP &Fast-SNN	&32/2	&50.72	&32	&41.97	&-8.75	&\textbf{3}\\
		\hline \hline
	\end{tabular}
\end{table*}

\begin{figure*}[t]
	\centering
	\includegraphics[width=\linewidth]{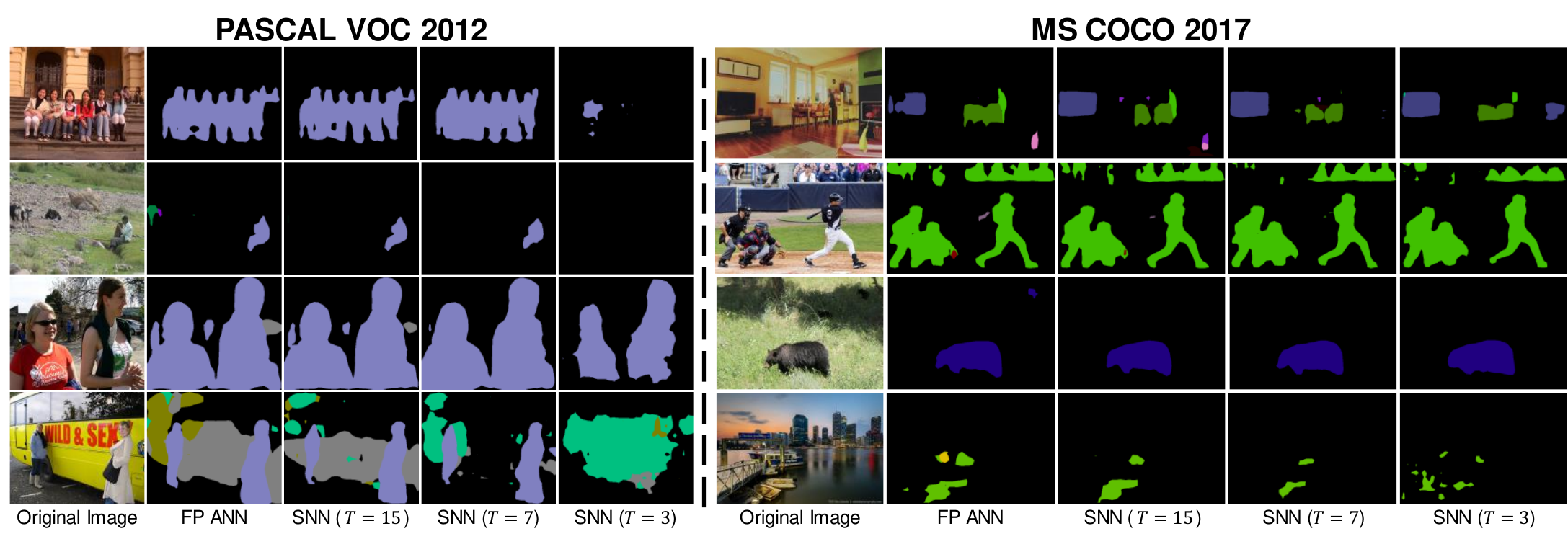}\\
	\caption{Visual quality comparison of semantic segmentation results on $val$ set of PASCAL VOC 2012 (left) and MS COCO 2017 (right) with the network architecture ResNet-34 + ASPP.  From left to right: the original image (input), results from full-precision (FP) ANN models, our SNN with $T=15, 7, 3$, respectively.}\label{fig:segmentation}
\end{figure*}

\subsection{Experimental Setup}
\ \indent
\textbf{Datasets.} We perform semantic segmentation task on two datasets: PASCAL VOC 2012 \cite{everingham2010pascal} and MS COCO 2017 \cite{lin2014microsoft}. PASCAL VOC 2012 is further augmented by the extra annotations provided by \cite{hariharan2011semantic}, resulting in 10,582 training images.

\textbf{Data preprocessing.} Following the original Deeplab protocol \cite{chen2014semantic,chen2017deeplab,chen2017rethinking}, we first standardize the data with the mean and standard deviation from the ImageNet dataset. Then we take a random $513 \times 513$ crop from the image. We further apply data augmentation by randomly scaling the input images (from 0.5 to 2.0) and flipping them horizontally at a chance of 50\%.  

\textbf{Network architecture.} We evaluate two Deeplab \cite{chen2017deeplab} based architectures for semantic segmentation. The first architecture is the VGG-9 defined in \cite{kim2022beyond}. To facilitate ANN-to-SNN conversion, we remove the average pooling layers and incorporate downsampling operations into convolution layers. With three downsampling layers of stride 2, the $output\_stride$ (defined as the ratio of input image spatial resolution to final output resolution \cite{chen2017rethinking}) of this VGG-9 architecture is 8. The second architecture comprises a backbone of ResNet-34 and an ASPP module. In the remainder of this paper, we refer to this architecture as ResNet-34 + ASPP. The ASPP module contains five parallel convolution layers: one $1\times1$ convolution and four $3\times3$ atrous convolutions with $rates=(6, 12, 18, 24)$ to capture multi-scale information, following the configurations in \cite{chen2017deeplab}. The $output\_stride$ of ResNet-34 + ASPP is 16.     

\textbf{Training details.} On PASCAL VOC and MS COCO, we fine-tune the models initialized from pre-trained ImageNet models to straightly adapt them to the semantic segmentation task. For VGG-9, we initialize the first seven convolution and batch normalization layers from the pre-trained VGG-16 in TorchVision \cite{torchvision2016}. For ResNet-34 + ASPP, we initialize the backbone from the pre-trained ResNet-34 in TorchVision \cite{torchvision2016}. Then we fine-tune the initialized models for 50 epochs using the standard SGD optimizer on both PASCAL VOC 2012 and MS COCO 2017. Following \cite{chen2017rethinking}, we initialize the learning rate to 0.007 and employ a `poly' learning rate policy where the initial learning rate is multiplied by  
\begin{equation}\label{eq:seg1} 
	(1 - \dfrac{iter}{max\_iter})^{power}
\end{equation}
with $power = 0.9$. We use a momentum of 0.9 and a weight decay of 1e-4. We progressively build ANNs with activations quantized to 4/3/2 bits and apply ANN-to-SNN conversion. 

\textbf{Evaluation metrics.} 
The performance is measured in terms of pixel intersection-over-union (IoU) averaged across the 21 classes (20 foreground object classes and one background class) on PASCAL VOC 2012 and 81 classes on MS COCO 2017 (80 foreground object classes and one background class).

\subsection{Overall Performance}
We summarize and compare the performance in Table \ref{tab:seg_results}. On PASCAL VOC 2012, we include the Spiking-Deeplab proposed by Kim et al. \cite{kim2022beyond} comparison. All numbers are taken from their paper. Compared with \cite{kim2022beyond}, our VGG-9 (latency is 3) outperforms their VGG-9 by 21.56\% mean IoU performance while using about $7\times$ fewer time steps. Compared with \cite{kim2022beyond}, our VGG-9 also achieves a limited performance gap between ANNs and SNNs. With a latency of 15, our converted VGG-9 even improves the ANN performance by 0.17\% at mean IoU. To further explore the capacity of deep SNNs in semantic segmentation, we apply our method to the more challenging ResNet-34 + ASPP. Our ResNet-34 + ASPP achieves 69.7\% mean IoU performance with a latency of 15, outperforming \cite{kim2022beyond} by 47.4\% mean IoU performance and 5 fewer time steps. For the first time, our method demonstrate that SNNs can achieve comparable performance to ANNs on the challenging MS COCO dataset performing semantic segmentation task. Comparing with corresponding ANNs, our VGG-9 has less than 2\% performance drop at mean IoU for all latency configurations (3,7,15). Our VGG-9 (latency is 15) achieves 31.14\% mean IoU performance, only 0.46\% lower than its ANN counterpart. For a deeper architecture, our ResNet-34 + ASPP (latency is 15) achieves 50.24\% mean IoU performance. We further provide visual results of our ResNet-34 + ASPP in Fig. \ref{fig:segmentation}. As shown in the figure, our SNNs are able to segment objects at a degree close to the full-precision ANN. For some objects (e.g., TV), our SNNs yield better results than the full-precision ANN.

\section{Conclusion}
In this work, we propose a framework to build a Fast-SNN with competitive performance (i.e., comparable with ANNs) and low inference latency (i.e., 3, 7, 15). Our basic idea is to minimize the quantization error and accumulating error. We show the equivalent mapping between temporal quantization in SNNs and spatial quantization in ANNs, based on which we transfer the minimization of the quantization error to quantized ANN training. This scheme facilitates ANN-to-SNN conversion by finding the optimal clipping range and the novel distributions of weights and activations for each layer. This mapping also makes the accumulating sequential error the only culprit of performance degradation when converting a quantized ANN to an SNN. To mitigate the impact of the sequential error at each layer, we propose a signed IF neuron to cancel the wrongly fired spikes. To alleviate the accumulating sequential error, we propose a layer-wise fine-tuning mechanism to minimize the difference between SNN firing rates and ANN activations. Our framework derives the upper bound of inference latency that guarantees no performance degradation when converting a quantized ANN to an SNN. Our method achieves state-of-the-art performance and low latency on various computer vision tasks, including image classification, object detection, and semantic segmentation. In addition, our SNNs that inherit the quantized weights in ANNs are intrinsically compatible with low-precision neuromorphic hardware.



\ifCLASSOPTIONcaptionsoff
\newpage
\fi

\bibliographystyle{IEEEtran}
\bibliography{IEEEabrv,Signy}

\end{document}